\documentclass[a4paper]{article}

\usepackage{graphicx}
\usepackage[margin=1in]{geometry}
\usepackage{float}
\usepackage{appendix}
\usepackage{tabularx}
\usepackage{booktabs}
\usepackage{natbib}
\usepackage{longtable}
\usepackage{ragged2e}
\usepackage{xurl}
\usepackage{rotating}
\usepackage{placeins}
\usepackage{hyperref}

\begin{document}

\title{EMIP: Embodied Multi-Party Interaction Platform}
\author{Koichiro Terao$^{1,2}$ \qquad Rikunari Sagara$^{2,3}$ \qquad Moe Nagao$^{1,2}$ \qquad Naoto Iwahashi$^{1,2}$\\ $^{1}$Okayama Prefectural University \quad $^{2}$AI and Humans Lab \quad $^{3}$University of Shizuoka
}
\date{}
\maketitle

\begin{abstract}
Experimental platforms that enable multiple participants to engage in spoken dialogue and perform physical actions concurrently in a shared environment are important for studying cooperation between humans and embodied AI agents. In this paper, we propose EMIP (Embodied Multi-Party Interaction Platform), a virtual environment platform built on Minecraft. Through an Internet-based client--server architecture, EMIP connects multiple humans and multiple AI agents to the same virtual environment and supports full-duplex speech communication and simultaneous physical interaction. It provides interfaces based on the Model Context Protocol (MCP) for environmental observation, physical action execution, and audio input and output, enabling researchers to develop their own agent control methods using large language models (LLMs). Additional features include centralized control of an agent group through a single EMIP agent client, collaborative object transportation by multiple agents, and experimental support functions such as recording and retrieving observation histories. By integrating these features into a reusable platform, EMIP aims to support research on human--human, human--agent, and agent--agent cooperation under different participant configurations and agent control methods. Usage examples illustrate concurrent speech and physical action, task coordination, and overlapping speech. The source code for the platform is publicly available on GitHub
(\url{https://github.com/AI-Humans-Lab/EMIP-Embodied-Multi-Party-Interaction-Platform}).
\end{abstract}

\begin{figure}[ht]
    \centering
    \includegraphics[width=.8\linewidth]{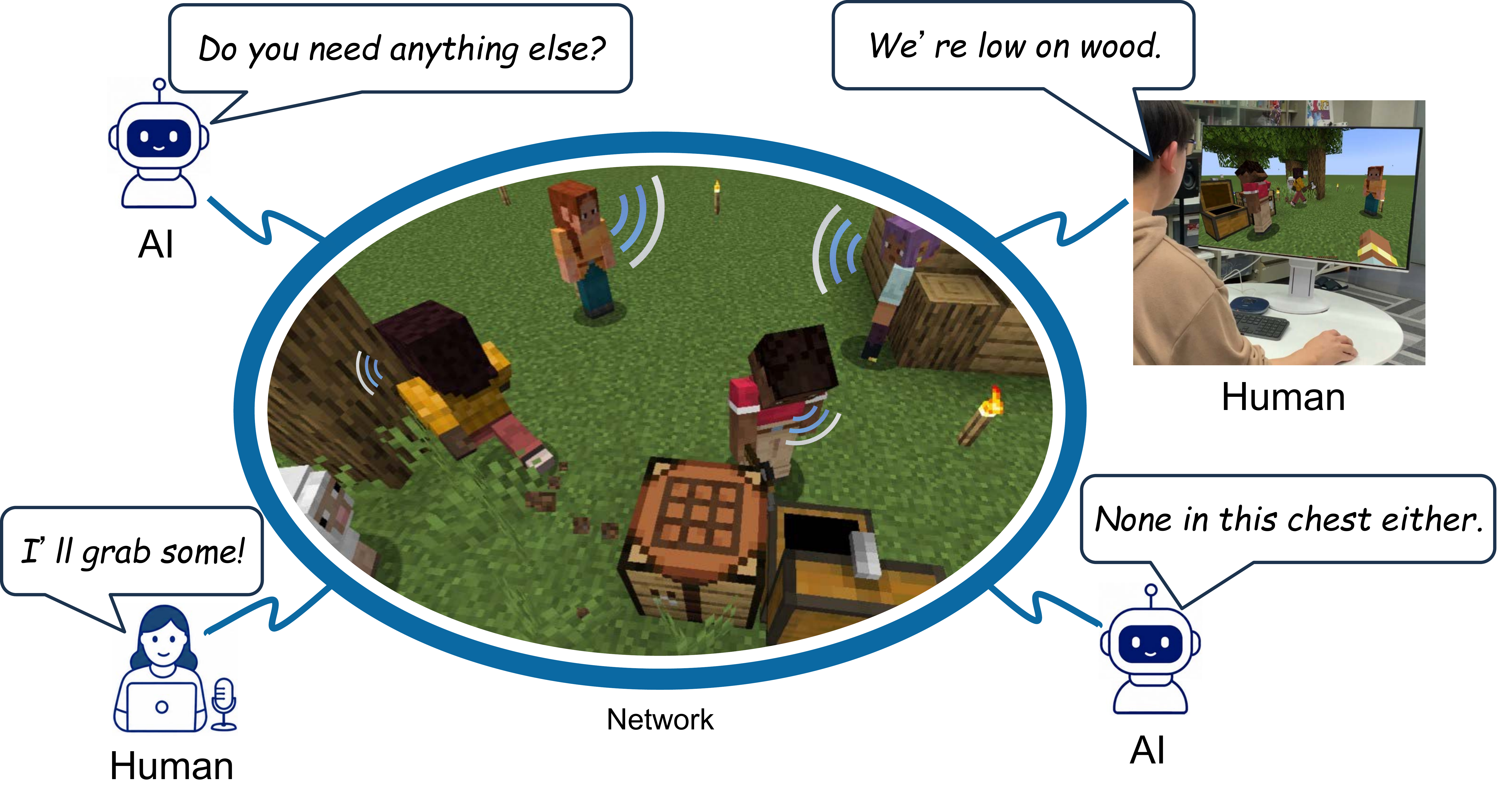}
    \caption{Multi-human and multi-agent interaction in EMIP.}
    \label{fig:emip_image}
\end{figure}

\section{Introduction}
\label{sec:introduction}

\begin{figure}[t]
    \centering
    \includegraphics[width=.75\linewidth]{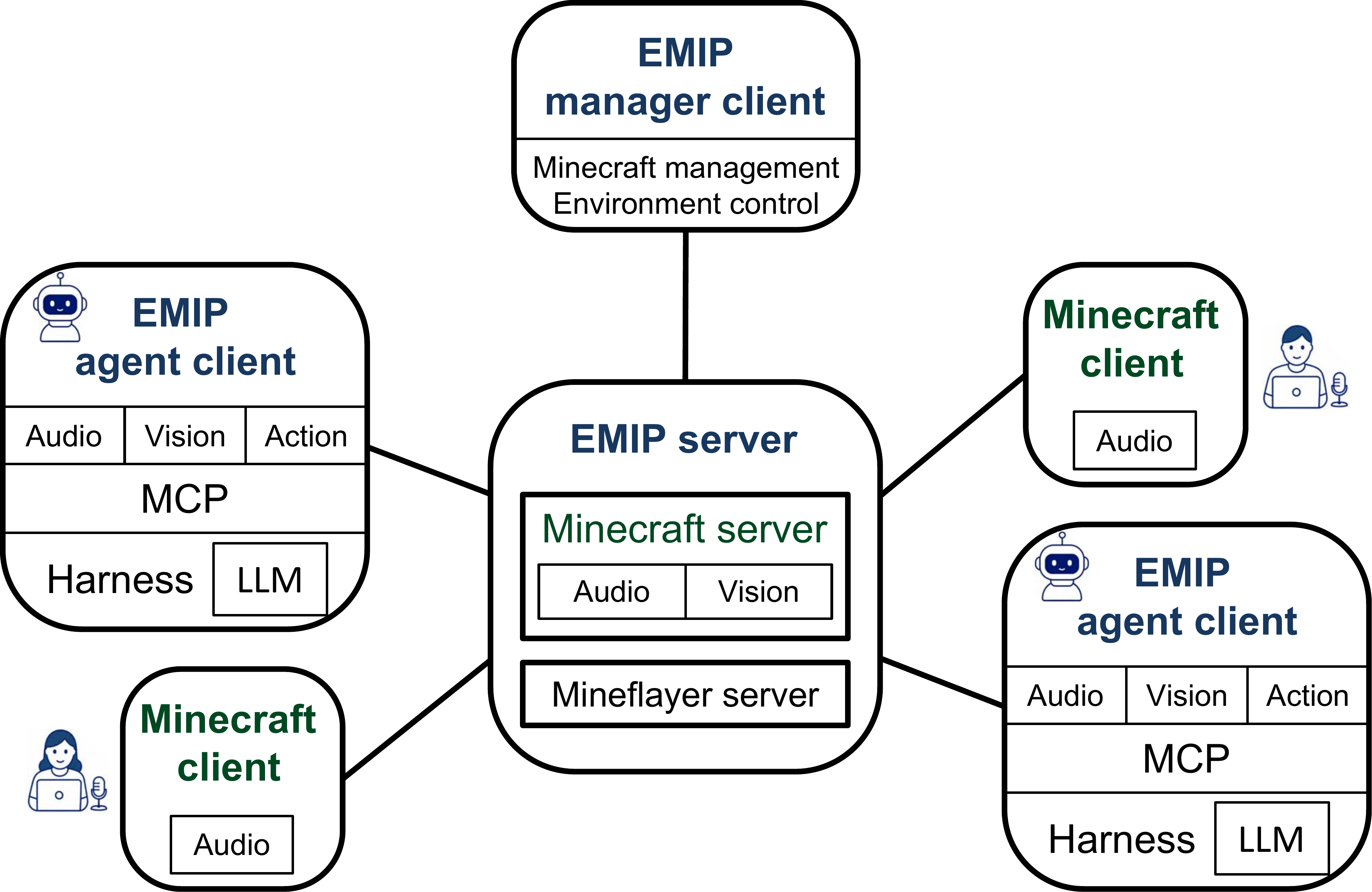}
    \caption{Software architecture of EMIP. Each EMIP agent client can centrally control a group of one or more embodied AI agents. By connecting multiple EMIP agent clients and human-operated Minecraft clients to the same EMIP server, the agent groups and human participants can interact in a shared virtual environment. The EMIP manager client manages the experimental environment separately from the control of agent groups. In the diagram, Vision denotes the function that provides structured information filtered by distance and occlusion.}
    \label{fig:software_configuration}
\end{figure}

The ability to share an environment with humans and perform tasks jointly through dialogue and physical actions is important for AI systems to function as cooperative partners in society. Such cooperation requires understanding a partner's intentions and situation, sharing information, allocating roles, and adjusting one's own behavior in response to others' utterances and actions. When multiple humans and embodied AI agents work together, dialogue and physical actions proceed concurrently, and each participant's behavior influences the decisions and actions of others. Understanding the mechanisms underlying such cooperation and developing and evaluating methods to support it require an experimental platform in which diverse participants can act and communicate in a shared environment.

Research on embodied AI and cooperation has produced virtual environment platforms, benchmarks for evaluating agent capabilities, and methods for cognition, planning, and control. Virtual environments provide settings in which environmental and task conditions can be specified and the resulting behavior and interactions can be observed. Benchmarks define evaluation tasks and metrics, while methods for cognition, planning, and control provide agents with the capabilities needed to perform those tasks. These interrelated efforts support research on agent capabilities and cooperative behavior.

Representative virtual environment platforms include Minecraft-based environments, Habitat, VirtualHome, ThreeDWorld, and Overcooked. Minecraft-based research environments support a wide variety of tasks, including exploration, survival, cooperation, and problem solving. Habitat supports the training and evaluation of embodied agents in photorealistic three-dimensional environments, including research on human--robot cooperation. VirtualHome represents everyday household activities as sequences of actions and interactions with objects and provides an environment in which virtual agents can perform these activities. ThreeDWorld integrates physics simulation with audiovisual data generation, supporting physical interactions involving objects and multiple agents. Overcooked-based research environments use simplified cooking tasks to investigate role allocation, action coordination, and adaptation to partners. Related studies using these platforms are summarized in Appendix~\ref{sec:related_work}.

Building on these studies, our aim is to provide a common experimental basis for investigating cooperation across different tasks, participant configurations, and agent control methods. Such an environment should allow researchers to vary the number and combination of human and AI participants and the agents' control methods while supporting spoken dialogue and physical actions that proceed concurrently. These capabilities are needed to examine how participants share information and coordinate their behavior during ongoing task execution under different experimental conditions.

To this end, we propose EMIP (Embodied Multi-Party Interaction Platform), a virtual environment platform built on Minecraft. EMIP is designed to accommodate most of the cooperative task types investigated in these related studies, including exploration, resource collection, construction, and tasks requiring role allocation and coordinated action. Through an Internet-based client--server architecture, it connects multiple humans and multiple AI agents in a shared environment and integrates full-duplex speech communication with simultaneous physical interaction. These capabilities allow participants to adjust their speech and actions in response to one another during joint tasks and provide a common setting for human--human, human--agent, and agent--agent cooperation.

EMIP provides interfaces based on the Model Context Protocol (MCP) for environmental observation, physical action execution, and audio input and output. Researchers can use these interfaces to develop their own agent control programs that use large language models (LLMs) to interpret the environment and other participants' situations, plan actions, and coordinate dialogue with physical actions. These interfaces allow researchers to configure experiments with different LLMs and methods for cognition, planning, and control while reusing the same underlying input and output functions. Observation histories can also be recorded and retrieved to support the analysis of interactions and cooperative processes.

EMIP also supports centralized control of a group of one or more embodied AI agents through a single EMIP agent client. Multiple EMIP agent clients can connect to the same EMIP server alongside human-operated Minecraft clients, allowing agent groups controlled by different clients and human participants to interact in a shared environment. This architecture supports experiments in which different control methods are assigned to different agent groups.

Additional features support physically coupled cooperative tasks through collaborative object transportation and provide experimental environment management through the EMIP manager client. The collaborative transportation capability allows multiple embodied agents, controlled by humans or AI, to move a shared object together. Because each agent's actions affect the object's motion, the participants must coordinate the direction and timing of their movements. Combined with spoken dialogue, this capability enables studies of how verbal coordination and physical actions are adjusted during collaborative transportation. The EMIP manager client supports experimental preparation, execution, and analysis through scene configuration, accelerated simulation, and recording and replay functions. Researchers can use these functions to set up task environments, adjust the rate at which the simulated environment advances, and record and review experimental sessions to examine participants' actions and interactions.

By integrating these capabilities into a common platform, EMIP enables researchers to extend established cooperative task paradigms to a broader range of multi-party interactions and to compare cooperative processes across participant configurations and agent control methods. This supports investigations of information sharing and role allocation, the coordination of speech and physical actions, and human acceptance and evaluation of agents. Providing these capabilities as reusable functions across tasks and experimental conditions reduces the effort required to build and configure experimental environments, allowing researchers to focus on the mechanisms of cooperation and the development and evaluation of agent control methods.

This paper describes the core and additional features and software architecture of EMIP. The functions of the MCP servers and the experimental environment management functions are detailed in Appendices~\ref{sec:agent_mcp} and~\ref{sec:manager_client}, respectively. Related studies are summarized in Appendix~\ref{sec:related_work}, and usage examples are presented in Appendix~\ref{sec:example_english}.

\section{Core Features}

EMIP is a research platform that enables multiple humans and embodied AI agents to interact through speech communication and physical actions in a shared virtual environment. It is intended to serve as a common platform that allows researchers to vary participant configurations and agent control methods according to their research objectives, without being restricted to a particular task or control algorithm. This section describes seven core features of EMIP.

\subsection{Multi-Human, Multi-Agent Interaction}

EMIP enables multiple humans and multiple AI agents to interact in the same Minecraft environment. It supports human--human, human--agent, and agent--agent cooperation within a common framework and provides an environment for studying multi-party interactions in cooperative tasks. Each participant performs physical actions in the shared environment while exchanging information through speech.

Researchers can configure diverse cooperative scenarios by varying the number of participants and the combination of humans and agents. Potential applications include experiments examining role allocation between humans and agents, assistance provided to humans by multiple agents, and information sharing within groups that include humans. The platform supports studies of both relationships between a single human and a single agent and situations in which relationships with multiple partners influence one another.

\subsection{Simultaneous Physical Interaction}

EMIP allows multiple participants to perform physical actions concurrently in a shared virtual environment, where the consequences of their actions affect other participants. Physical actions and speech communication can also proceed concurrently, providing an environment for studying dialogue during task execution and how participants adjust their actions in response to a partner's utterances and actions.

For example, research tasks can involve participants sharing updates on their progress while dividing responsibility for exploration and resource collection, or changing their assigned tasks in response to the status of a partner's work.

\subsection{Full-Duplex Speech Communication}

EMIP supports full-duplex speech communication, allowing audio transmission and reception to occur simultaneously. This feature serves as communication infrastructure for studying dialogue involving overlapping speech, backchannel responses, and interruptions. Participants can receive audio from other participants while speaking, enabling spoken interactions that do not require strictly alternating periods of speaking and listening. Speech communication proceeds concurrently with physical actions, allowing researchers to configure cooperative scenarios in which participants exchange information while working.

Combining this feature with simultaneous physical interaction enables researchers to analyze the temporal relationship between speech and physical actions and how they are coordinated, rather than evaluating each modality in isolation.

\subsection{Internet-Based Client-Server Architecture}

EMIP employs a client--server architecture that connects participants to the virtual environment over the Internet. As shown in Figure~\ref{fig:software_configuration}, both Minecraft clients and EMIP agent clients connect to an EMIP server that provides the shared environment. Human participants control their avatars through Minecraft clients, while AI agents connect through EMIP agent clients. The server side comprises a Minecraft server and a Mineflayer server, allowing human-controlled agents and AI agents to act in the same virtual environment. The EMIP manager client provides scene configuration, accelerated simulation, and recording and replay functions to support experimental setup, execution, and analysis.

\subsection{Agent Model Context Protocol (MCP) Servers}

EMIP provides MCP interfaces, collectively referred to as Agent MCP Servers, through which agent control programs interact with the Minecraft environment and participate in spoken dialogue. The Agent MCP Servers comprise the Observation-Action MCP server, Voice MCP server, and Belief MCP server. The Observation-Action MCP server generates and provides observations from an agent's subjective perspective and supports action execution in the environment. The Voice MCP server provides access to received audio data and speech recognition results and supports text-to-speech synthesis and the transmission of synthesized audio to Minecraft. The Belief MCP server manages accumulated subjective states and provides access to observation histories.

Researchers can connect their own agent control programs to these MCP interfaces and investigate different models and methods while reusing the underlying input and output functions. Details of the resources and tools exposed by each MCP server are provided in Appendix~\ref{sec:agent_mcp}.

\subsection{Modular Configuration}

EMIP can be used in a minimal configuration consisting of Minecraft clients connected to an EMIP server. In this configuration, human participants control agents in Minecraft. Adding EMIP agent clients connects AI agents to the environment, while adding an EMIP manager client enables experimental environment management. This modular structure allows researchers to add the clients required for their research objectives.

\subsection{Software Portability}

To simplify the setup and recreation of experimental environments, EMIP packages its main software components and their dependencies in Docker containers. This reduces dependence on the host operating system and the effort required for researchers to install and configure runtime environments and libraries individually. Using the same container configuration across computers also facilitates sharing and recreating experimental environments.

\section{Additional Features}

This section describes features for centralized control of agent groups, collaborative object transportation, experimental support, and long-term execution.

\subsection{Group-Level Centralized Control}

EMIP supports centralized control of a group of one or more embodied AI agents through a single EMIP agent client. Here, a group refers to the set of agents controlled by the same EMIP agent client. The EMIP agent client centralizes connection and execution management for the agents in the group. The scope of information shared among agents for decision-making is determined by the researcher's control program. This feature allows researchers to develop control programs for role allocation and action coordination within a group without requiring a separate client for each agent. Connecting multiple EMIP agent clients to the same EMIP server also allows agent groups controlled by different clients to interact in a shared virtual environment.

\subsection{Collaborative Object Transportation}

EMIP provides action capabilities that allow multiple embodied agents to transport the same object collaboratively. These capabilities enable cooperative tasks in which each agent manipulates the object and coordinates its movements with those of the other agents to transport it to a shared destination. Because each agent's actions affect the motion of the shared object, the agents must coordinate aspects of their movements such as direction and timing. This enables joint actions in which agents interact mechanically through the shared object in the virtual environment. This feature supports collaborative transportation both among AI agents and between human-controlled agents and AI agents.

Combining these action capabilities with spoken dialogue allows researchers to configure cooperative scenarios in which verbal coordination and the physical actions involved in transporting an object proceed concurrently. This provides an experimental setting for investigating how agreements reached through dialogue are reflected in physical actions and how dialogue and actions are adjusted as the situation changes during transportation.

\subsection{Experimental Support Functions}

To support the preparation, execution, and analysis of experiments on cooperation, EMIP provides scene configuration, accelerated simulation, and recording and replay functions through the EMIP manager client. Scene configuration allows researchers to set up experimental environments tailored to their research tasks, while recording and replay allow them to review experimental sessions and analyze them retrospectively. The accelerated simulation feature and its constraints are described in Section~\ref{sec:long-term-execution}. Together, these functions reduce the effort required throughout the experimental workflow, from setting up experimental conditions and running experiments to reviewing and analyzing the results.

The Observation-Action MCP server and Belief MCP server also provide information management and retrieval functions to support analyses based on each agent's observation history and subjective state. The Observation-Action MCP server manages each agent's subjective observations along with their associated tick values, which indicate logical time in Minecraft. The Belief MCP server provides functions to retrieve stored observations and events within a specified tick range and to reconstruct an agent's subjective state at a specified time from those records. After an experiment, researchers can use these functions to examine what information each agent observed and when, as well as the subjective state maintained by the Belief MCP server on the basis of those records. This supports analysis of cooperative processes and differences in agents' understanding of the environment.

\subsection{Long-Term Execution} 
\label{sec:long-term-execution}

EMIP is designed to support experiments and continuous simulations over extended periods. These experiments can address phenomena that unfold over longer time scales, such as sustained interactions, environmental changes, the accumulation of experience by agents, and the development of relationships among them.

To facilitate such studies, EMIP provides an accelerated simulation feature that allows researchers to configure the rate at which the Minecraft environment advances and the observation interval. When agent control processes can keep pace with the advancing environment, this feature can reduce the real time required to simulate a given period of activity.

However, LLM inference times, audio processing times, and human participants' response times impose constraints that are separate from the environment's progression rate. The speedup achieved for an experiment as a whole therefore depends on the participant configuration, agent control method, and computing environment.

\section{System Architecture}

Figure~\ref{fig:software_configuration} shows the software architecture of EMIP. EMIP comprises an EMIP server that provides the shared virtual environment, Minecraft clients operated by human participants, EMIP agent clients that handle AI agent input, output, and control, and an EMIP manager client that manages the experimental environment. These components communicate over a network, separating environment management, human participation, and AI agent control while allowing all participants to interact within a shared environment.

\subsection{EMIP Server}

The EMIP server consists of a Minecraft server and a Mineflayer server. The Minecraft server manages the state of the shared virtual environment and updates that state in response to the actions of human participants and AI agents. It also provides the Audio and Vision functions shown in Figure~\ref{fig:software_configuration}.

The Mineflayer server acts as an intermediary between agent control programs running in EMIP agent clients and the Minecraft environment. It manages each AI agent's connection to Minecraft and executes action requests received from the control programs to control the agents in the environment.

\subsection{Minecraft Client}

Human participants use Minecraft clients to connect to the EMIP server and perform tasks jointly with other participants through their avatars' physical actions and spoken communication. An extension for audio input and output enables speech communication with other humans and AI agents. Connecting multiple Minecraft clients allows multiple humans to participate in the same environment.

\subsection{EMIP Agent Client}

The EMIP agent client provides AI agents with access to the Minecraft environment and handles their input, output, and control. It supports full-duplex audio communication, allowing agents to receive other participants' audio while speaking.

A single EMIP agent client can centrally manage and control a group of one or more agents. Thus, the agents embodied in the virtual environment are distinct from the software client that controls them, and a separate client is not required for each agent.

In the configuration shown in Figure~\ref{fig:software_configuration}, the EMIP agent client comprises Audio, Vision, and Action components, MCP interfaces, a Harness, and an LLM. The MCP interfaces connect agent control programs to the environment input and output functions.

The Observation-Action MCP server handles environment observation and physical action execution, the Voice MCP server handles audio input and output, and the Belief MCP server manages subjective states and observation histories and provides access to them. Each agent is assigned a separate instance of each of these three types of MCP server. Researchers use these common interfaces to construct agent control programs.

\subsubsection*{Observation-Action MCP}

The Observation-Action MCP server provides environment observations and executes physical actions. It filters information received from Minecraft based on distance from the agent and occlusion by blocks, producing structured observations from the agent's subjective perspective. The Vision component in Figure~\ref{fig:software_configuration} denotes the function that provides these observations.

The control program retrieves the latest observation in JSON format through an MCP resource and obtains queued observations in chronological order through a tool. Each observation is associated with a tick value indicating logical time in Minecraft.

Physical actions are executed through a tool exposed by the same server. This tool runs JavaScript code on the Mineflayer server to control the corresponding agent in the Minecraft environment.

The server also supports starting and stopping observation processing and waiting for new observations. These functions allow the control program to adjust the timing of observation retrieval and processing in relation to action execution. Details of each tool are provided in Appendix~\ref{sec:agent_mcp}.

\subsubsection*{Voice MCP}

The Voice MCP server provides functions for configuring audio communication, retrieving received audio data and speech recognition results, and transmitting synthesized speech or externally generated audio to Minecraft. It exposes these functions through MCP tools and does not expose resources.

On the receiving side, the server supports retrieval of all stored audio data or speech recognition results, as well as incremental retrieval of data added since the previous incremental retrieval. On the transmitting side, it synthesizes speech from specified text using a specified voice and sends the resulting audio to Minecraft. It can also stream audio generated by an external dialogue model.

Concurrent audio reception and transmission allow agents to continue receiving other participants' audio while speaking. This provides a basis for full-duplex spoken dialogue with overlapping utterances. When combined with a compatible dialogue model, these functions support interactions in which a human participant gives backchannels or additional instructions during an agent's utterance. Using audio input and output concurrently with physical actions also allows agents to receive and transmit speech while performing tasks.

\subsubsection*{Belief MCP}

The Belief MCP server manages each agent's accumulated subjective state and observation history. Here, a subjective state is a representation of the environment maintained by an agent on the basis of its own observations, including information about agents, blocks, and containers. When the Harness registers an observation obtained from the Observation-Action MCP server, the subjective state is updated and the observation is added to the history. Because the subjective state is based on registered observations, it does not necessarily match the actual state of the environment.

The control program retrieves a snapshot containing the subjective state and observation history in JSON format through an MCP resource. It also uses tools to retrieve observations and events within a specified tick range and to reconstruct a past subjective state from information stored at or before a specified tick. The program can therefore access previously acquired information and reconstruct the subjective state associated with a particular point in the experiment.

Through the Belief MCP server, the control program can also access estimates of other agents' subjective states and observation histories by specifying the target agent's name. These estimates are produced by the Harness from the controlled agent's own observations; they are not obtained by directly accessing other agents' internal states or histories. This allows the control program to use both the information available to the controlled agent and its estimates of the information available to others.

\subsubsection*{Agent Control with the Harness and LLM}

In the configuration shown in Figure~\ref{fig:software_configuration}, the Harness links the LLM to environment observations, physical actions, and audio input and output through MCP interfaces. The Harness and LLM together handle agent cognition, planning, and dialogue control, while the MCP servers provide access to the functions needed to interact with the Minecraft environment.

This separation allows researchers to develop their own agent control programs while reusing the platform's environment input and output functions. The connected control program determines how to interpret incoming speech, when to respond, and whether to interrupt speech output. These decisions are distinct from the mechanisms that receive and transmit audio.

\subsection{EMIP Manager Client}

The EMIP manager client provides an interface for managing the experimental environment independently of agent control. It connects to the EMIP server through the environment management functions shown in Figure~\ref{fig:software_configuration} and supports scene configuration, adjustment of the environment's progression rate and observation interval, and replay and review of recorded experimental sessions to examine participants' actions and interactions. Details of these functions are provided in Appendix~\ref{sec:manager_client}.

Centralized management and control of agent groups are handled by EMIP agent clients. This division allows researchers to manage the virtual environment and control the agents operating within it as separate parts of the experimental system.

\section{Conclusion}

This paper presented EMIP, a Minecraft-based virtual environment platform for studying multi-party cooperation among humans and embodied AI agents, and described its design principles, core features, and software architecture. EMIP enables multiple humans and AI agents to participate in a shared environment and interact through spoken dialogue and physical actions. By combining full-duplex speech communication with simultaneous physical interaction, it supports cooperative tasks in which dialogue and physical actions proceed concurrently.

The architecture separates the components responsible for environment management, human participation, and AI agent control and connects them over a network. MCP interfaces provide access to environmental observations, physical action execution, and audio input and output, allowing researchers to develop and connect their own agent control programs. A single EMIP agent client can centrally manage and control a group of one or more agents, providing a basis for developing control methods for individual agent behavior, role allocation, and action coordination within agent groups.

The main contribution of this study is the integration of communication, environment input and output, agent control interfaces, and experimental support functions into a reusable platform for research on multi-party cooperation. This platform is intended to support the development of methods for cognition, planning, and control in cooperative settings, as well as studies of the relationship between dialogue and physical actions, role allocation among participants, and human acceptance and evaluation of AI agents.

Future work includes evaluating the platform's utility across diverse cooperative tasks and participant configurations, assessing communication and execution performance with multiple participants connected simultaneously, and extending the functions for experimental configuration and analysis. We also plan to improve setup documentation and usage examples to help researchers use and extend EMIP for their own research objectives.

\bibliographystyle{acl_natbib_noin}
\bibliography{related_work}

@inproceedings{Gan2020ThreeDWorld,
 author = {Gan, Chuang and Schwartz, Jeremy and Alter, Seth and Mrowca, Damian and Schrimpf, Martin and Traer, James and De Freitas, Julian and Kubilius, Jonas and Bhandwaldar, Abhishek and Haber, Nick and Sano, Megumi and Kim, Kuno and Wang, Elias and Lingelbach, Michael and Curtis, Aidan and Feigelis, Kevin and Bear, Daniel and Gutfreund, Dan and Cox, David and Torralba, Antonio and DiCarlo, James J and Tenenbaum, Josh and McDermott, Josh and Yamins, Dan},
 booktitle = {The Annual Conference on Neural Information Processing Systems Track on Datasets and Benchmarks},
 editor = {J. Vanschoren and S. Yeung},
 pages = {},
 title = {{ThreeDWorld}: A Platform for Interactive Multi-Modal Physical Simulation},
 url = {https://datasets-benchmarks-proceedings.neurips.cc/paper_files/paper/2021/file/735b90b4568125ed6c3f678819b6e058-Paper-round1.pdf},
 volume = {1},
 year = {2021}
}

@inproceedings{Zhang2025COMBO,
 author = {Zhang, Hongxin and Wang, Zeyuan and Lyu, Qiushi and Zhang, Zheyuan and Chen, Sunli and Shu, Tianmin and Dariush, Behzad and Lee, Kwonjoon and Du, Yilun and Gan, Chuang},
 booktitle = {International Conference on Learning Representations},
 editor = {Y. Yue and A. Garg and N. Peng and F. Sha and R. Yu},
 pages = {49996--50019},
 title = {{COMBO}: Compositional World Models for Embodied Multi-Agent Cooperation},
 url = {https://proceedings.iclr.cc/paper_files/paper/2025/file/7d03c6bf9f07acb4038eea96c63db52d-Paper-Conference.pdf},
 volume = {2025},
 year = {2025}
}

@inproceedings{Puig2020Watch-and-help,
title={Watch-And-Help: A Challenge for Social Perception and Human-{AI} Collaboration},
author={Xavier Puig and Tianmin Shu and Shuang Li and Zilin Wang and Yuan-Hong Liao and Joshua B. Tenenbaum and Sanja Fidler and Antonio Torralba},
booktitle={International Conference on Learning Representations},
year={2021},
url={https://openreview.net/forum?id=w_7JMpGZRh0}
}

@INPROCEEDINGS{Puig2023NOPA,
  author={Puig, Xavier and Shu, Tianmin and Tenenbaum, Joshua B. and Torralba, Antonio},
  booktitle={2023 IEEE International Conference on Robotics and Automation}, 
  title={{NOPA}: Neurally-guided Online Probabilistic Assistance for Building Socially Intelligent Home Assistants}, 
  year={2023},
  volume={},
  number={},
  pages={7628-7634},
  doi={10.1109/ICRA48891.2023.10161352}}

@inproceedings{Zhang2024Building,
 author = {Zhang, Hongxin and Du, Weihua and Shan, Jiaming and Zhou, Qinhong and Du, Yilun and Tenenbaum, Joshua B and Shu, Tianmin and Gan, Chuang},
 booktitle = {International Conference on Learning Representations},
 editor = {B. Kim and Y. Yue and S. Chaudhuri and K. Fragkiadaki and M. Khan and Y. Sun},
 pages = {19373--19401},
 title = {Building Cooperative Embodied Agents Modularly with Large Language Models},
 url = {https://proceedings.iclr.cc/paper_files/paper/2024/file/54b8b4e0b4ba4aad112e84f32e3b5dbb-Paper-Conference.pdf},
 volume = {2024},
 year = {2024}
}

@INPROCEEDINGS{Ying2024GOMA,
  title     = "{GOMA}: Proactive embodied cooperative communication via
               goal-oriented mental alignment",
  author    = "Ying, Lance and Jha, Kunal and Aarya, Shivam and Tenenbaum,
               Joshua B and Torralba, Antonio and Shu, Tianmin",
  booktitle = "2024 IEEE/RSJ International Conference on Intelligent Robots and
               Systems (IROS)",
  pages     = "7099--7106",
  month     =  oct,
  year      =  2024,
  language  = "en"
}

@ARTICLE{Guo2026Embodied,
  title     = "Embodied {LLM} agents learn to cooperate in organized teams",
  author    = "Guo, Xudong and Huang, Kaixuan and Liu, Jiale and Fan, Wenhui and
               V{\'e}lez, Natalia and Wu, Qingyun and Wang, Huazheng and Griffiths,
               Thomas L and Wang, Mengdi",
  journal   = "IEEE Transactions on Computational Social Systems",
  publisher = "Institute of Electrical and Electronics Engineers (IEEE)",
  volume    =  13,
  number    =  2,
  pages     = "2514--2530",
  month     =  apr,
  year      =  2026,
  language  = "en"
}

@ARTICLE{Seo2025REVECA,
  title     = "{REVECA}: Adaptive Planning and Trajectory-based Validation in
               cooperative language agents using information {RElevance} and
               relative Proximity",
  author    = "Seo, SeungWon and Noh, SeongRae and Lee, Junhyeok and Lim, SooBin
               and Lee, Won Hee and Kang, HyeongYeop",
  journal   = "Proc. Conf. AAAI Artif. Intell.",
  publisher = "Association for the Advancement of Artificial Intelligence (AAAI)",
  volume    =  39,
  number    =  22,
  pages     = "23295--23303",
  month     =  apr,
  year      =  2025,
  language  = "en"
}

@inproceedings{Liu2025CaPo,
 author = {Liu, Jie and Zhou, Pan and Du, Yingjun and Tan, Ah-Hwee and Snoek, Cees G and Sonke, Jan-jakob and Gavves, Efstratios},
 booktitle = {International Conference on Learning Representations},
 editor = {Y. Yue and A. Garg and N. Peng and F. Sha and R. Yu},
 pages = {70791--70815},
 title = {{CaPo}: Cooperative Plan Optimization for Efficient Embodied Multi-Agent Cooperation},
 url = {https://proceedings.iclr.cc/paper_files/paper/2025/file/b07091c16719ad3990e3d1ccee6641f1-Paper-Conference.pdf},
 volume = {2025},
 year = {2025}
}

@misc{Li2025Learn,
      title={Learn as Individuals, Evolve as a Team: Multi-agent {LLMs} Adaptation in Embodied Environments}, 
      author={Xinran Li and Chenjia Bai and Zijian Li and Jiakun Zheng and Ting Xiao and Jun Zhang},
      year={2025},
      eprint={2506.07232},
      archivePrefix={arXiv},
      primaryClass={cs.MA},
      url={https://arxiv.org/abs/2506.07232}, 
}

@InProceedings{Zu2025Collaborative,
    author    = {Zu, Lizheng and Lin, Lin and Fu, Song and Zhao, Na and Zhou, Pan},
    title     = {Collaborative Tree Search for Enhancing Embodied Multi-Agent Collaboration},
    booktitle = {The IEEE/CVF Conference on Computer Vision and Pattern Recognition},
    month     = {June},
    year      = {2025},
    pages     = {29513-29522}
}

@misc{Wang2025Collaborative,
      title={Collaborative Belief Reasoning with {LLMs} for Efficient Multi-Agent Collaboration}, 
      author={Zhimin Wang and Duo Wu and Shaokang He and Jinghe Wang and Linjia Kang and Jing Yu and Kai Zhu and Jiawei Li and Zhi Wang},
      year={2025},
      eprint={2509.21981},
      archivePrefix={arXiv},
      primaryClass={cs.AI},
      url={https://arxiv.org/abs/2509.21981}, 
}

@ARTICLE{Jeon2025Mind-Map,
  title     = "{Mind-Map Agent}: Enhancing Cooperative Task Planning through
               Communication Alignment with Large Language Models",
  author    = "Jeon, Hobeom and Kim, Hyungmin and Kim, Dohyung and Jang, Minsu
               and Kim, Jaehong",
  year      =  2025,
  journal   = "The Annual Conference on Neural Information Processing Systems Workshop on Bridging Language, Agent, and World Models",
  pages     = "1--12",
  month     =  dec,
  language  = "en"
}

@inproceedings{Suprabha2026Improving,
  title     = "Improving cooperation in collaborative embodied {AI}",
  author    = "Suprabha, Hima Jacob Leven and Nagesh, Laxmi Nag Laxminarayan and
               Nair, Ajith and Selvaster, Alvin Reuben Amal and Khan, Ayan and
               Damarla, Raghuram and Samuel, Sanju Hannah and Perumal, Sreenithi
               Saravana and Puech, Titouan and Marella, Venkataramireddy and
               Sonar, Vishal and Suglia, Alessandro and Lemon, Oliver",
  year      =  2025,
  booktitle = "UK Workshop on Computational Intelligence",
  pages     = "145--157",
  language  = "en"
}

@inproceedings{Seo2026From,
  title={From Assumptions to Actions: Turning {LLM} Reasoning into Uncertainty-Aware Planning for Embodied Agents},
  author={SeungWon Seo and SooBin Lim and Seongrae Noh and Haneul Kim and HyeongYeop Kang},
  booktitle={The Fourteenth International Conference on Learning Representations},
  year={2026},
  url={https://openreview.net/forum?id=GODFBZhFcX}
}

@INPROCEEDINGS{Zhang2026MiTa,
  title     = "{MiTa}: A hierarchical multi-agent collaboration framework with
               memory-integrated and task allocation",
  author    = "Zhang, Xiaojie and Wu, Jianhan and Qu, Xiaoyang and Wang,
               Jianzong",
  booktitle = "IEEE International Conference on Acoustics,
               Speech and Signal Processing",
  pages     = "20586--20590",
  month     =  may,
  year      =  2026,
  language  = "en"
}

@ARTICLE{Yu2026Co-NavGPT,
  author={Yu, Bangguo and Yuan, Qihao and Li, Kailai and Kasaei, Hamidreza and Cao, Ming},
  journal={IEEE Robotics and Automation Letters}, 
  title={Co-{NavGPT}: Multirobot Cooperative Visual Semantic Navigation Using Vision Language Models}, 
  year={2026},
  volume={11},
  number={2},
  pages={2122-2129},
  doi={10.1109/LRA.2025.3645650}}

@INCOLLECTION{Yu2022Learning,
  title     = "Learning efficient multi-agent cooperative visual exploration",
  author    = "Yu, Chao and Yang, Xinyi and Gao, Jiaxuan and Yang, Huazhong and
               Wang, Yu and Wu, Yi",
  booktitle = "Computer Vision -- ECCV 2022",
  publisher = "Springer Nature Switzerland",
  pages     = "497--515",
  series    = "Lecture Notes in Computer Science",
  year      =  2022,
  language  = "en"
}

@inproceedings{Yu2023Asynchronous,
author = {Yu, Chao and Yang, Xinyi and Gao, Jiaxuan and Chen, Jiayu and Li, Yunfei and Liu, Jijia and Xiang, Yunfei and Huang, Ruixin and Yang, Huazhong and Wu, Yi and Wang, Yu},
title = {Asynchronous Multi-Agent Reinforcement Learning for Efficient Real-Time Multi-Robot Cooperative Exploration},
year = {2023},
isbn = {9781450394321},
booktitle = {The 2023 International Conference on Autonomous Agents and Multiagent Systems},
pages = {1107–1115},
numpages = {9},
location = {London, United Kingdom},
series = {AAMAS '23}
}

@InProceedings{Szot2023Adaptive,
  title = 	 {Adaptive Coordination in Social Embodied Rearrangement},
  author =       {Szot, Andrew and Jain, Unnat and Batra, Dhruv and Kira, Zsolt and Desai, Ruta and Rai, Akshara},
  booktitle = 	 {International Conference on Machine Learning PMLR},
  pages = 	 {33365--33380},
  year = 	 {2023},
  editor = 	 {Krause, Andreas and Brunskill, Emma and Cho, Kyunghyun and Engelhardt, Barbara and Sabato, Sivan and Scarlett, Jonathan},
  volume = 	 {202},
  month = 	 {23--29 Jul},
  url = 	 {https://proceedings.mlr.press/v202/szot23a.html}
}

@inproceedings{Puig2024Habitat,
 author = {Puig, Xavier and Undersander, Eric and Szot, Andrew and Dallaire Cote, Mikael and Yang, Tsung-Yen and Partsey, Ruslan and Desai, Ruta and Clegg, Alexander and Hlavac, Michal and Min, So Yeon and Vondru\v{s}, Vladim\'{\i}r and Gervet, Theophile and Berges, Vincent-Pierre and Turner, John and Maksymets, Oleksandr and Kira, Zsolt and Kalakrishnan, Mrinal and Malik, Jitendra and Chaplot, Devendra and Jain, Unnat and Batra, Dhruv and Rai, Akshara and Mottaghi, Roozbeh},
 booktitle = {International Conference on Learning Representations},
 editor = {B. Kim and Y. Yue and S. Chaudhuri and K. Fragkiadaki and M. Khan and Y. Sun},
 pages = {15306--15336},
 title = {Habitat 3.0: A Co-Habitat for Humans, Avatars, and Robots},
 url = {https://proceedings.iclr.cc/paper_files/paper/2024/file/430894999584d0bd358611e2ecf00b15-Paper-Conference.pdf},
 volume = {2024},
 year = {2024}
}

@ARTICLE{Shen2025Enhancing,
  title     = "Enhancing multi-robot semantic navigation through Multimodal
               Chain-of-Thought score collaboration",
  author    = "Shen, Zhixuan and Luo, Haonan and Chen, Kexun and Lv, Fengmao and
               Li, Tianrui",
  journal   = "Proc. Conf. AAAI Artif. Intell.",
  publisher = "Association for the Advancement of Artificial Intelligence (AAAI)",
  volume    =  39,
  number    =  14,
  pages     = "14664--14672",
  month     =  apr,
  year      =  2025,
  language  = "en"
}

@inproceedings{Scofano2025Following,
 author = {Scofano, Luca and Sampieri, Alessio and Campari, Tommaso and Sacco, Valentino and Spinelli, Indro and Ballan, Lamberto and Galasso, Fabio},
 booktitle = {International Conference on Learning Representations},
 editor = {Y. Yue and A. Garg and N. Peng and F. Sha and R. Yu},
 pages = {47674--47694},
 title = {Following the Human Thread in Social Navigation},
 url = {https://proceedings.iclr.cc/paper_files/paper/2025/file/76b010c4d21f97b7b113383ce7160cbc-Paper-Conference.pdf},
 volume = {2025},
 year = {2025}
}

@inproceedings{Chen2025EMOS,
 author = {Chen, Junting and Yu, Checheng and Zhou, Xunzhe and Xu, Tianqi and Mu, Yao and Hu, Mengkang and Shao, Wenqi and Wang, Yikai and Li, Guohao and Shao, Lin},
 booktitle = {International Conference on Learning Representations},
 editor = {Y. Yue and A. Garg and N. Peng and F. Sha and R. Yu},
 pages = {56670--56690},
 title = {{EMOS}: Embodiment-aware Heterogeneous Multi-robot Operating System with {LLM} Agents},
 url = {https://proceedings.iclr.cc/paper_files/paper/2025/file/8ec46cecdae5407662899c5f6698cd8b-Paper-Conference.pdf},
 volume = {2025},
 year = {2025}
}

@inproceedings{Chang2025PARTNR,
 author = {Chang, Matthew and Chhablani, Gunjan and Clegg, Alexander and Dallaire Cote, Mikael and Desai, Ruta and Hlavac, Michal and Karashchuk, Vladimir and Krantz, Jacob and Mottaghi, Roozbeh and Parashar, Priyam and Patki, Siddharth and Prasad, Ishita and Puig, Xavier and Rai, Akshara and Ramrakhya, Ram and Tran, Daniel and Truong, Joanne and Turner, John and Undersander, Eric and Yang, Tsung-Yen},
 booktitle = {International Conference on Learning Representations},
 editor = {Y. Yue and A. Garg and N. Peng and F. Sha and R. Yu},
 pages = {65205--65268},
 title = {{PARTNR}: A Benchmark for Planning and Reasoning in Embodied Multi-agent Tasks},
 url = {https://proceedings.iclr.cc/paper_files/paper/2025/file/a3cf318fbeec1126da21e9185ae9908c-Paper-Conference.pdf},
 volume = {2025},
 year = {2025}
}

@inproceedings{Jeon2025Communicative,
title={Communicative {PARTNR}: Natural Language Communication under Partial Observability in Human{\textendash}Robot Collaboration},
author={HoBeom Jeon and Hyungmin Kim and Dohyung Kim and Minsu Jang and Jaehong Kim},
booktitle={ICCV Workshop on Human-Robot-Scene Interaction and Collaboration},
year={2025},
url={https://openreview.net/forum?id=r1f58OTBbY}
}

@inproceedings{Wang2025CoNavBench,
  title={{CoNavBench}: Collaborative Long-Horizon Vision-Language Navigation Benchmark},
  author={Wang, Tianhang and Li, Xinhai and Lu, Fan and Gong, Tianshi and Dong, Jiankun and Xue, Weiyi and Qu, Sanqing and Bai, Chenjia and Chen, Guang},
  booktitle={The Fourteenth International Conference on Learning Representations},
  year={2026},
  url={https://openreview.net/forum?id=bMrH2PFMsi}
}

@INPROCEEDINGS{Ma2025COOPERA,
  title     = "{COOPERA}: Continual open-ended human-robot assistance",
  author    = "Ma, Chenyang and Lu, Kai and Desai, Ruta and Puig, Xavier and
               Markham, Andrew and Trigoni, Niki",
  editor    = "Belgrave, D and Zhang, C and Lin, H and Pascanu, R and Koniusz, P
               and Ghassemi, M and Chen, N",
  year      =  2025,
  booktitle = "The Annual Conference on Neural Information Processing Systems",
  volume    = "38",
  pages     = "38557--38591",
}

@misc{Juneja2026EnactToM,
      title={{EnactToM}: An Evolving Benchmark for Functional Theory of Mind in Embodied Agents}, 
      author={Gurusha Juneja and Dylan Lu and Saaket Agashe and Parth Diwane and Edward Gunn and Jayanth Srinivasa and Gaowen Liu and William Yang Wang and Yali Du and Xin Eric Wang},
      year={2026},
      eprint={2605.09826},
      archivePrefix={arXiv},
      primaryClass={cs.AI},
      url={https://arxiv.org/abs/2605.09826}, 
}

@INPROCEEDINGS{Liu2026Vulcan,
  author={Liu, Shengding and Yan, Qiben},
  booktitle={IEEE Conference on Computer Communications Workshop on Embodied Intelligence Networks},
  title={Vulcan: Vision-Language-Model Enhanced Multi-Agent Cooperative Navigation for Indoor Fire-Disaster Response},
  year={2026},
  volume={},
  number={},
  pages={1-6},
  doi={10.1109/INFOCOM59046.2026.11680073}
}

@misc{Peng2026Seeing,
      title={Seeing Together: Multi-Robot Cooperative Egocentric Spatial Reasoning with Multimodal Large Language Models}, 
      author={Kunyu Peng and Zhikun Zhou and Kailun Yang and Di Wen and Ruiping Liu and Yufan Chen and Junwei Zheng and Hao Shi and Yi Zhou and M. Saquib Sarfraz and Danda Pani Paudel and Luc Van Gool},
      year={2026},
      eprint={2605.18431},
      archivePrefix={arXiv},
      primaryClass={cs.CV},
      url={https://arxiv.org/abs/2605.18431}, 
}

@article{Demir2020Understanding,
title = {Understanding human-robot teams in light of all-human teams: Aspects of team interaction and shared cognition},
journal = {International Journal of Human-Computer Studies},
volume = {140},
pages = {102436},
year = {2020},
issn = {1071-5819},
doi = {https://doi.org/10.1016/j.ijhcs.2020.102436},
url = {https://www.sciencedirect.com/science/article/pii/S1071581920300380},
author = {Mustafa Demir and Nathan J. McNeese and Nancy J. Cooke}
}

@inproceedings{Johnson2016The,
  title     = {The {Malmo} Platform for Artificial Intelligence Experimentation},
  author    = {Johnson, Matthew and Hofmann, Katja and Hutton, Tim and Bignell, David},
  booktitle = {The Twenty-Fifth International Joint Conference on Artificial Intelligence, {IJCAI-16}},
  editor    = {Subbarao Kambhampati},
  pages     = {4246--4247},
  year      = {2016},
  month     = {7},
  url       = {https://ijcai.org/proceedings/2016/643},
}

@ARTICLE{Callaghan2016Investigating,
  title     = "Investigating the role of {Minecraft} in educational learning
               environments",
  author    = "Callaghan, Noelene",
  journal   = "Educational Media International",
  publisher = "Informa UK Limited",
  volume    =  53,
  number    =  4,
  pages     = "244--260",
  month     =  oct,
  year      =  2016,
  language  = "en"
}

@inproceedings{Perez-Liebana2019The,
author={ Perez-Liebana, Diego and Hofmann, Katja and Mohanty, Sharada Prasanna and Kuno, Noburu and Kramer, Andre and Devlin, Sam and Gaina, Raluca D and Ionita, Daniel },
title={ The Multi-Agent Reinforcement Learning in {Malm{\"O}} ({MARL{\"O}}) Competition },
year={ 2018 },
booktitle={ Challenges in Machine Learning (CiML; NIPS Workshop) },
pages={ 1--4 }
}

@misc{Gray2019CraftAssist,
      title={{CraftAssist}: A Framework for Dialogue-enabled Interactive Agents}, 
      author={Jonathan Gray and Kavya Srinet and Yacine Jernite and Haonan Yu and Zhuoyuan Chen and Demi Guo and Siddharth Goyal and C. Lawrence Zitnick and Arthur Szlam},
      year={2019},
      eprint={1907.08584},
      archivePrefix={arXiv},
      primaryClass={cs.AI},
      url={https://arxiv.org/abs/1907.08584}, 
}

@inproceedings{Narayan-Chen2019Collaborative,
    title = "Collaborative Dialogue in {M}inecraft",
    author = "Narayan-Chen, Anjali  and
      Jayannavar, Prashant  and
      Hockenmaier, Julia",
    editor = "Korhonen, Anna  and
      Traum, David  and
      M{\`a}rquez, Llu{\'i}s",
    booktitle = "The 57th Annual Meeting of the {A}ssociation for {C}omputational {L}inguistics",
    month = jul,
    year = "2019",
    url = "https://aclanthology.org/P19-1537/",
    doi = "10.18653/v1/P19-1537",
    pages = "5405--5415"
}

@inproceedings{Bara2021MindCraft,
    title = "{M}ind{C}raft: Theory of Mind Modeling for Situated Dialogue in Collaborative Tasks",
    author = "Bara, Cristian-Paul  and
      CH-Wang, Sky  and
      Chai, Joyce",
    editor = "Moens, Marie-Francine  and
      Huang, Xuanjing  and
      Specia, Lucia  and
      Yih, Scott Wen-tau",
    booktitle = "The 2021 Conference on Empirical Methods in Natural Language Processing",
    month = nov,
    year = "2021",
    url = "https://aclanthology.org/2021.emnlp-main.85/",
    doi = "10.18653/v1/2021.emnlp-main.85",
    pages = "1112--1125"
}

@inproceedings{Paleja2022The,
author = {Paleja, Rohan and Ghuy, Muyleng and Arachchige, Nadun R. and Jensen, Reed and Gombolay, Matthew},
title = {The utility of explainable {AI} in ad hoc human-machine teaming},
year = {2021},
isbn = {9781713845393},
booktitle = {International Conference on Neural Information Processing Systems},
articleno = {47},
numpages = {14}
}

@INPROCEEDINGS{Ichikawa2022Analysis,
  title     = "Analysis of Dialogue in Human-Human Collaboration in {Minecraft}",
  author    = "Ichikawa, Takuma and Higashinaka, Ryuichiro",
  booktitle = "The Thirteenth Language Resources and Evaluation
               Conference",
  pages     = "4051--4059",
  year      =  2022
}

@INPROCEEDINGS{Volum2022Craft,
  title     = "Craft an Iron Sword: Dynamically Generating Interactive Game
               Characters by Prompting Large Language Models Tuned on Code",
  author    = "Volum, Ryan and Rao, Sudha and Xu, Michael and DesGarennes,
               Gabriel and Brockett, Chris and Van Durme, Benjamin and Deng,
               Olivia and Malhotra, Akanksha and Dolan, William B",
  booktitle = "The Wordplay: When Language Meets Games
               Workshop",
  pages     = "25--43",
  year      =  2022
}

@InProceedings{Kiseleva2022Interactive,
  title = 	 {Interactive Grounded Language Understanding  in a Collaborative Environment: {IGLU} 2021},
  author =       {Kiseleva, Julia and Li, Ziming and Aliannejadi, Mohammad and Mohanty, Shrestha and ter Hoeve, Maartje and Burtsev, Mikhail and Skrynnik, Alexey and Zholus, Artem and Panov, Aleksandr and Srinet, Kavya and Szlam, Arthur and Sun, Yuxuan and Hofmann, Katja and C{\^o}t{\'e}, Marc-Alexandre and Awadallah, Ahmed and Abdrazakov, Linar and Churin, Igor and Manggala, Putra and Naszadi, Kata and van der Meer, Michiel and Kim, Taewoon},
  booktitle = 	 {The Annual Conference on Neural Information Processing Systems Competitions and Demonstrations Track PMLR},
  pages = 	 {146--161},
  year = 	 {2022},
  editor = 	 {Kiela, Douwe and Ciccone, Marco and Caputo, Barbara},
  volume = 	 {176},
  month = 	 {06--14 Dec},
  url = 	 {https://proceedings.mlr.press/v176/kiseleva22a.html}
}

@inproceedings{Ichikawa2023Modeling,
    title = "Modeling Collaborative Dialogue in {M}inecraft with Action-Utterance Model",
    author = "Ichikawa, Takuma  and
      Higashinaka, Ryuichiro",
    editor = "Li, Dongfang  and
      Mahendra, Rahmad  and
      Tang, Zilu Peter  and
      Jang, Hyeju  and
      Murawaki, Yugo  and
      Wong, Derek Fai",
    booktitle = "The 13th International Joint Conference on Natural Language Processing and the 3rd Conference of the {A}sia-Pacific Chapter of the {A}ssociation for {C}omputational {L}inguistics: Student Research Workshop",
    month = nov,
    year = "2023",
    publisher = "Association for Computational Linguistics",
    url = "https://aclanthology.org/2023.ijcnlp-srw.10/",
    doi = "10.18653/v1/2023.ijcnlp-srw.10",
    pages = "75--81"
}

@INPROCEEDINGS{Yocum2023Mitigating,
  title     = "Mitigating generative agent social dilemmas",
  author    = "Yocum, Julian and Christoffersen, Phillip J K and Damani, Mehul
               and Svegliato, Justin and Hadfield-Menell, Dylan and Russell,
               Stuart",
  booktitle = "NeurIPS 2023 Foundation Models for Decision Making Workshop",
  year      =  2023
}

@ARTICLE{Deffner2024Collective,
  title     = "Collective incentives reduce over-exploitation of social
               information in unconstrained human groups",
  author    = "Deffner, Dominik and Mezey, David and Kahl, Benjamin and
               Schakowski, Alexander and Romanczuk, Pawel and Wu, Charley M and
               Kurvers, Ralf H J M",
  journal   = "Nat. Commun.",
  publisher = "Nature Publishing Group",
  volume    =  15,
  number    =  1,
  pages     =  2683,
  month     =  mar,
  year      =  2024,
  language  = "en"
}

@misc{Yu2024MineLand,
      title={{MineLand}: Simulating Large-Scale Multi-Agent Interactions with Limited Multimodal Senses and Physical Needs}, 
      author={Xianhao Yu and Jiaqi Fu and Renjia Deng and Wenjuan Han},
      year={2024},
      eprint={2403.19267},
      archivePrefix={arXiv},
      primaryClass={cs.CL},
      url={https://arxiv.org/abs/2403.19267}, 
}

@inproceedings{Gong2024MindAgent,
    title = "{M}ind{A}gent: Emergent Gaming Interaction",
    author = "Gong, Ran  and
      Huang, Qiuyuan  and
      Ma, Xiaojian  and
      Noda, Yusuke  and
      Durante, Zane  and
      Zheng, Zilong  and
      Terzopoulos, Demetri  and
      Fei-Fei, Li  and
      Gao, Jianfeng  and
      Vo, Hoi",
    editor = "Duh, Kevin  and
      Gomez, Helena  and
      Bethard, Steven",
    booktitle = "Findings of the {A}ssociation for {C}omputational {L}inguistics: {NAACL} 2024",
    month = jun,
    year = "2024",
    url = "https://aclanthology.org/2024.findings-naacl.200/",
    doi = "10.18653/v1/2024.findings-naacl.200",
    pages = "3154--3183"
}

@inproceedings{Dong2024VillagerAgent,
    title = "{V}illager{A}gent: A Graph-Based Multi-Agent Framework for Coordinating Complex Task Dependencies in {M}inecraft",
    author = "Dong, Yubo  and
      Zhu, Xukun  and
      Pan, Zhengzhe  and
      Zhu, Linchao  and
      Yang, Yi",
    editor = "Ku, Lun-Wei  and
      Martins, Andre  and
      Srikumar, Vivek",
    booktitle = "Findings of the {A}ssociation for {C}omputational {L}inguistics: {ACL} 2024",
    month = aug,
    year = "2024",
    publisher = "Association for Computational Linguistics",
    url = "https://aclanthology.org/2024.findings-acl.964/",
    doi = "10.18653/v1/2024.findings-acl.964",
    pages = "16290--16314"
}

@INPROCEEDINGS{Rao2024Collaborative,
  title     = "Collaborative Quest Completion with {LLM}-driven Non-Player
               Characters in {Minecraft}",
  author    = "Rao, Sudha and Xu, Weijia and Xu, Michael and Leandro, Jorge J G
               and Lobb, Ken and DesGarennes, Gabriel and Brockett, Chris and
               Dolan, Bill",
  booktitle = "Wordplay: When Language Meets Games Workshop at ACL",
  month     =  aug,
  year      =  2024,
  language  = "en"
}

@misc{Altera2024Project,
      title={{Project Sid}: Many-agent simulations toward {AI} civilization}, 
      author={{Altera.AL} and Andrew Ahn and Nic Becker and Stephanie Carroll and Nico Christie and Manuel Cortes and Arda Demirci and Melissa Du and Frankie Li and Shuying Luo and Peter Y Wang and Mathew Willows and Feitong Yang and Guangyu Robert Yang},
      year={2024},
      eprint={2411.00114},
      archivePrefix={arXiv},
      primaryClass={cs.AI},
      url={https://arxiv.org/abs/2411.00114}, 
}

@misc{Long2024TeamCraft,
      title={{TeamCraft}: A Benchmark for Multi-Modal Multi-Agent Systems in {Minecraft}}, 
      author={Qian Long and Zhi Li and Ran Gong and Ying Nian Wu and Demetri Terzopoulos and Xiaofeng Gao},
      year={2024},
      eprint={2412.05255},
      archivePrefix={arXiv},
      primaryClass={cs.AI},
      url={https://arxiv.org/abs/2412.05255}, 
}

@ARTICLE{Bendell2025Artificial,
  title     = "Artificial social intelligence in teamwork: how team traits
               influence human-{AI} dynamics in complex tasks",
  author    = "Bendell, Rhyse and Williams, Jessica and Fiore, Stephen M and
               Jentsch, Florian",
  journal   = "Front. Robot. AI",
  publisher = "Frontiers Media SA",
  volume    =  12,
  pages     =  1487883,
  month     =  feb,
  year      =  2025,
  language  = "en"
}

@ARTICLE{Wu2025Adaptive,
  title     = "Adaptive mechanisms of social and asocial learning in immersive
               collective foraging",
  author    = "Wu, Charley M and Deffner, Dominik and Kahl, Benjamin and Meder,
               Bj{\"o}rn and Ho, Mark K and Kurvers, Ralf H J M",
  journal   = "Nature Communications",
  publisher = "Springer Science and Business Media LLC",
  volume    =  16,
  number    =  1,
  pages     =  3539,
  month     =  apr,
  year      =  2025,
  language  = "en"
}

@inproceedings{Pyarelal2025MultiCAT,
    title = "{M}ulti{CAT}: Multimodal Communication Annotations for Teams",
    author = "Pyarelal, Adarsh  and
      Culnan, John M  and
      Qamar, Ayesha  and
      Krishnaswamy, Meghavarshini  and
      Wang, Yuwei  and
      Jeong, Cheonkam  and
      Chen, Chen  and
      Miah, Md Messal Monem  and
      Hormozi, Shahriar  and
      Tong, Jonathan  and
      Huang, Ruihong",
    editor = "Chiruzzo, Luis  and
      Ritter, Alan  and
      Wang, Lu",
    booktitle = "Findings of the {A}ssociation for {C}omputational {L}inguistics: {NAACL} 2025",
    month = apr,
    year = "2025",
    url = "https://aclanthology.org/2025.findings-naacl.61/",
    doi = "10.18653/v1/2025.findings-naacl.61",
    pages = "1077--1111",
    ISBN = "979-8-89176-195-7"
}

@misc{Sagara2025BeliefNest,
      title={{BeliefNest}: A Joint Action Simulator for Embodied Agents with Theory of Mind}, 
      author={Rikunari Sagara and Koichiro Terao and Naoto Iwahashi},
      year={2025},
      eprint={2505.12321},
      archivePrefix={arXiv},
      primaryClass={cs.AI},
      url={https://arxiv.org/abs/2505.12321}, 
}

@INPROCEEDINGS{Schipper2025PillagerBench,
  title     = "{PillagerBench}: Benchmarking {LLM}-Based Agents in Competitive
               {Minecraft} Team Environments",
  author    = "Schipper, Olivier and Zhang, Yudi and Du, Yali and Pechenizkiy,
               Mykola and Fang, Meng",
  booktitle = "IEEE Conference on Games",
  pages     = "1--15",
  month     =  aug,
  year      =  2025,
  language  = "en"
}

@ARTICLE{Coulon2026Acquiring,
  title     = "Acquiring complex knowledge and skills through digital
               simulation-based training: Evidence from an agile project
               management teaching experience",
  author    = "Coulon, Thibaut and Cheikh-Ammar, Mustapha and Bourdeau, Simon
               and Petit, Marie-Claude",
  journal   = "Br. J. Educ. Technol.",
  publisher = "Wiley",
  volume    =  57,
  number    =  1,
  pages     = "243--271",
  month     =  Aug,
  year      =  2025,
  language  = "en"
}

@ARTICLE{Willett2025Building,
  title     = "Building an {LLM}-based teammate in {Minecraft}: A testbed for
               human-{AI} collaboration",
  author    = "Willett, Matthew M and Cohen, Myke C and Huang, Lixiao and Teo,
               Aaron and Klinefelter, Zachary and Bautista, Peter and Rajib,
               Rijvi and Reynolds, Jonathan and Fouse, Adam and Cooke, Nancy J",
  journal   = "Proc. Hum. Factors Ergon. Soc. Annu. Meet.",
  publisher = "SAGE Publications",
  volume    =  69,
  number    =  1,
  pages     = "1713--1715",
  month     =  sep,
  year      =  2025,
  language  = "en"
}

@inproceedings{Hirota2025Investigating,
    title = "Investigating Feasibility of Large Language Model Agent Collaboration in {M}inecraft and Comparison with Human-Human Collaboration",
    author = "Hirota, Yuki  and
      Higashinaka, Ryuichiro",
    editor = "Inui, Kentaro  and
      Sakti, Sakriani  and
      Wang, Haofen  and
      Wong, Derek F.  and
      Bhattacharyya, Pushpak  and
      Banerjee, Biplab  and
      Ekbal, Asif  and
      Chakraborty, Tanmoy  and
      Singh, Dhirendra Pratap",
    booktitle = "The 14th International Joint Conference on Natural Language Processing and the 4th Conference of the {A}sia-Pacific Chapter of the {A}ssociation for {C}omputational {L}inguistics",
    month = dec,
    year = "2025",
    url = "https://aclanthology.org/2025.ijcnlp-long.73/",
    doi = "10.18653/v1/2025.ijcnlp-long.73",
    pages = "1333--1347",
    ISBN = "979-8-89176-298-5"
}

@ARTICLE{Han2026Improving,
  title     = "Improving the efficiency of collaboration between humans and
               embodied {AI} agents in {3D} virtual environments",
  author    = "Han, Seowon and Lee, Kang Hoon",
  journal   = "Applied Sciences",
  publisher = "MDPI AG",
  volume    =  16,
  number    =  2,
  pages     =  1135,
  month     =  jan,
  year      =  2026,
  language  = "en"
}

@misc{Wang2026Requesting,
      title={Requesting Expert Reasoning: Augmenting {LLM} Agents with Learned Collaborative Intervention}, 
      author={Zhiming Wang and Jinwei He and Feng Lu},
      year={2026},
      eprint={2602.22546},
      archivePrefix={arXiv},
      primaryClass={cs.AI},
      url={https://arxiv.org/abs/2602.22546}, 
}

@INPROCEEDINGS{Yamazaki2026Wont,
  title     = "Won't, not can't: Designing a refusal agent for long-term
               human–{AI} collaboration",
  author    = "Yamazaki, Takanori and Kadowaki, Rio and Muto, Takeshi and Muto,
               Yumiko",
  year      =  2026,
  booktitle = "The Extended Abstracts of the 2026 CHI Conference
               on Human Factors in Computing Systems",
  pages     = "1--5",
  month     =  apr,
  language  = "en"
}

@INPROCEEDINGS{Ying2026Pragmatic,
  title     = "Pragmatic embodied spoken instruction following in human-robot
               collaboration with theory of mind",
  author    = "Ying, Lance and Li, Xinyi and Aarya, Shivam and Fang, Yizirui and
               Liu, Jason Xinyu and Yin, Yifan and Tellex, Stefanie and
               Tenenbaum, Joshua B and Shu, Tianmin",
  booktitle = "IEEE International Conference on Robotics and Automation",
  pages     = "6502--6509",
  month     =  jun,
  year      =  2026,
  language  = "en"
}

@INPROCEEDINGS{Fan2025SoMi-ToM,
  title     = "{SoMi}-{ToM}: Evaluating multi-perspective theory of mind in
               embodied social interactions",
  author    = "Fan, Xianzhe and Zhou, Xuhui and Jin, Chuanyang and Nottingham,
               Kolby and Zhu, Hao and Sap, Maarten",
  editor    = "Belgrave, D and Zhang, C and Lin, H and Pascanu, R and Koniusz, P
               and Ghassemi, M and Chen, N",
  year      =  2025,
  booktitle = "The Annual Conference on Neural Information Processing Systems",
  volume    = "38",
  pages     = "82621--82652",
}

@inproceedings{Zhang2026CommCP,
    title={{CommCP}: Efficient Multi-Agent Coordination via {LLM}-Based Communication with Conformal Prediction},
    author={Zhang, Xiaopan and Wang, Zejin and Li, Zhixu and Yao, Jianpeng and Li, Jiachen},
    booktitle={IEEE International Conference on Robotics and Automation},
    pages = {8613-8620},
    year={2026}
}

@inproceedings{Yu2026SurveilNav,
  author       = {Ming{-}Ming Yu and
                  Qunbo Wang and
                  Rongtao Xu and
                  Yanghong Mei and
                  Yirong Yang and
                  Longteng Guo and
                  Wenjun Wu and
                  Jing Liu},
  title        = {{SurveilNav}: Collaborative Object Goal Navigation with Robot and Surveillance
                  System},
  booktitle    = {{IEEE} International Conference on Robotics and Automation},
  pages        = {14671--14678},
  year         = {2026},
  url          = {https://doi.org/10.1109/ICRA57385.2026.11697410},
  doi          = {10.1109/ICRA57385.2026.11697410},
  bibsource    = {dblp computer science bibliography, https://dblp.org}
}

@INPROCEEDINGS{Zhao2026DRAMA,
  title     = "{DRAMA}: Next-Gen Dynamic Orchestration for Resilient Multi-Agent
               Ecosystems in Flux",
  author    = "Zhao, Xinkui and Zhang, Yifan and Liu, Sai and Wang, Naibo and
               Cheng, Guanjie and Xu, Yueshen and Liu, Chang and Deng, Shuiguang
               and Yin, Jianwei",
  booktitle = "The IEEE/CVF Conference on Computer Vision and
               Pattern Recognition",
  pages     = "1020--1030",
  year      =  2026
}

@INPROCEEDINGS{Zhang2026MindPower,
  title     = "{MindPower}: Enabling Theory-of-Mind Reasoning in {VLM}-based
               Embodied Agents",
  author    = "Zhang, Ruoxuan and Zheng, Qiyun and Zhou, Zhiyu and Liao, Ziqi
               and Wu, Siyu and Jiang-Lin, Jian-Yu and Wen, Bin and Xie, Hongxia
               and Fu, Jianlong and Cheng, Wen-Huang",
  booktitle = "The IEEE/CVF Conference on Computer Vision and
               Pattern Recognition",
  pages     = "29631--29641",
  year      =  2026
}

@ARTICLE{Dodampegama2026Collaborate,
AUTHOR={Dodampegama, Hasra  and Sridharan, Mohan },
TITLE={Collaborate and explain on-the-fly: knowledge-based reasoning and learning in ad hoc teamwork},
JOURNAL={Frontiers in Artificial Intelligence},
VOLUME={Volume 9 - 2026},
YEAR={2026},
URL={https://www.frontiersin.org/journals/artificial-intelligence/articles/10.3389/frai.2026.1765191},
DOI={10.3389/frai.2026.1765191},
ISSN={2624-8212}}

@misc{Liu2026MECoBench,
      title={{MECoBench}: A Systematic Study of Multimodal Agent Collaboration in Embodied Environments}, 
      author={Qingyun Liu and Jiwen Zhang and Jingyi Hu and Siyuan Wang and Zhongyu Wei},
      year={2026},
      eprint={2606.31966},
      archivePrefix={arXiv},
      primaryClass={cs.MA},
      url={https://arxiv.org/abs/2606.31966}, 
}

@inproceedings{Dongre2026Embodied,
    title = "Embodied Multi-Agent Coordination by Aligning World Models Through Dialogue",
    author = "Dongre, Vardhan  and
      Hakkani-Tur, Dilek",
    editor = "Choi, Jinho D.  and
      Chen, Yun-Nung  and
      Funakoshi, Kotaro  and
      Emami, Ali",
    booktitle = "The 27th Annual Meeting of the Special Interest Group on Discourse and Dialogue",
    month = aug,
    year = "2026",
    url = "https://aclanthology.org/2026.sigdial-1.21/",
    doi = "10.18653/v1/2026.sigdial-1.21",
    pages = "293--306"
}

@inproceedings{Sacco2026Robots,
  title     = {Robots Ask the Way: Communication-Enabled Social Navigation},
  author    = {Valentino Sacco and Luca Scofano and Indro Spinelli and Fabio Galasso},
  booktitle = {IEEE/RSJ International Conference on Intelligent Robots and Systems},
  year      = {2026},
}

@InProceedings{Qian2026CollabBench,
  title = 	 {{C}ollab{B}ench: Benchmarking and Unleashing Collaborative Ability of {LLM}s with Diverse Players via Proactive Engagement},
  author =       {Qian, Hong and Liu, Yuanhao and Zhou, Zihan and Zhang, Zongbao and Ge, Hanjie and Shi, Haotian and Dou, Liang and Wang, Xiangfeng and Yang, Jing-Wen and Zhou, Aimin},
  booktitle = 	 {International Conference on Machine Learning PMLR},
  pages = 	 {100787--100848},
  year = 	 {2026},
  editor = 	 {Zhang, Tong and Dudik, Miroslav and Jaggi, Martin and Agarwal, Alekh and Li, Sharon and Schuurmans, Dale and Zhu, Jerry and Berkenkamp, Felix and Dong, Hanze and Bietti, Alberto},
  volume = 	 {306},
  month = 	 {06--11 Jul},
  url = 	 {https://proceedings.mlr.press/v306/qian26c.html}
}

@InProceedings{Zhou2026LLawCo,
  title = 	 {{LL}aw{C}o: Learning Laws of Cooperation for Modeling Embodied Multi-Agent Behavior},
  author =       {Zhou, Qinhong and Gan, Chuang and Cherian, Anoop},
  booktitle = 	 {International Conference on Machine Learning PMLR},
  pages = 	 {164849--164866},
  year = 	 {2026},
  editor = 	 {Zhang, Tong and Dudik, Miroslav and Jaggi, Martin and Agarwal, Alekh and Li, Sharon and Schuurmans, Dale and Zhu, Jerry and Berkenkamp, Felix and Dong, Hanze and Bietti, Alberto},
  volume = 	 {306},
  month = 	 {06--11 Jul},
  url = 	 {https://proceedings.mlr.press/v306/zhou26i.html}
}

@inproceedings{Ahn2026Lies,
  author = {Jaewoo Ahn and Junseo Kim and Hyunseo Kim and Heeseung Yun and Jaehyeon Son and Zsolt Kira and Gunhee Kim},
  title = {{Lies We Can See: Joint Verbal and Non-Verbal Deception by VLM Agents in Embodied Social Interactions}},
  booktitle = {Conference on Language Modeling The First Workshop on Agent Behavior},
  year = {2026},
  url = {https://openreview.net/forum?id=tFdiQLpHJp},
}

@inproceedings{Chai2025CausalMACE,
    title = "{C}ausal{MACE}: Causality Empowered Multi-Agents in {M}inecraft Cooperative Tasks",
    author = "Chai, Qi  and
      Zheng, Zhang  and
      Ren, Junlong  and
      Ye, Deheng  and
      Lin, Zichuan  and
      Wang, Hao",
    editor = "Christodoulopoulos, Christos  and
      Chakraborty, Tanmoy  and
      Rose, Carolyn  and
      Peng, Violet",
    booktitle = "Findings of the {A}ssociation for {C}omputational {L}inguistics: {EMNLP} 2025",
    month = nov,
    year = "2025",
    url = "https://aclanthology.org/2025.findings-emnlp.777/",
    doi = "10.18653/v1/2025.findings-emnlp.777",
    pages = "14410--14426",
    ISBN = "979-8-89176-335-7"
}

@misc{White2025Collaborating,
      title={Collaborating Action by Action: A Multi-agent {LLM} Framework for Embodied Reasoning}, 
      author={Isadora White and Kolby Nottingham and Ayush Maniar and Max Robinson and Hansen Lillemark and Mehul Maheshwari and Lianhui Qin and Prithviraj Ammanabrolu},
      year={2025},
      eprint={2504.17950},
      archivePrefix={arXiv},
      primaryClass={cs.MA},
      url={https://arxiv.org/abs/2504.17950}, 
}

@INPROCEEDINGS{Lica2025MindForge,
  title     = "{MindForge}: Empowering embodied agents with theory of mind for
               lifelong cultural learning",
  author    = "Lic{\u{a}}, Mircea and Shirekar, Ojas and Colle, Baptiste and Raman,
               Chirag",
  editor    = "Belgrave, D and Zhang, C and Lin, H and Pascanu, R and Koniusz, P
               and Ghassemi, M and Chen, N",
  year      =  2025,
  booktitle = "The Annual Conference on Neural Information Processing Systems",
  volume    = "38",
  pages     = "188292--188321"
}

@misc{Jian2026Gated,
      title={Gated Coordination for Efficient Multi-Agent Collaboration in {Minecraft} Game}, 
      author={HuaDong Jian and Chenghao Li and Haoyu Wang and Jiajia Shuai and Jinyu Guo and Yang Yang and Chaoning Zhang},
      year={2026},
      eprint={2604.18975},
      archivePrefix={arXiv},
      primaryClass={cs.MA},
      url={https://arxiv.org/abs/2604.18975}, 
}

@misc{Yi2026Multi-agent,
      title={Multi-agent Framework for Time-Sensitive Complementary Collaboration in {Minecraft}}, 
      author={Juheon Yi and Jinglu Wang and Xiaoyi Zhang and Yan Lu},
      year={2026},
      eprint={2606.15684},
      archivePrefix={arXiv},
      primaryClass={cs.AI},
      url={https://arxiv.org/abs/2606.15684}, 
}

@inproceedings{Zhang2025Leveraging,
    title = "Leveraging Dual Process Theory in Language Agent Framework for Real-time Simultaneous Human-{AI} Collaboration",
    author = "Zhang, Shao  and
      Wang, Xihuai  and
      Zhang, Wenhao  and
      Li, Chaoran  and
      Song, Junru  and
      Li, Tingyu  and
      Qiu, Lin  and
      Cao, Xuezhi  and
      Cai, Xunliang  and
      Yao, Wen  and
      Zhang, Weinan  and
      Wang, Xinbing  and
      Wen, Ying",
    editor = "Che, Wanxiang  and
      Nabende, Joyce  and
      Shutova, Ekaterina  and
      Pilehvar, Mohammad Taher",
    booktitle = "The 63rd Annual Meeting of the {A}ssociation for {C}omputational {L}inguistics (Volume 1: Long Papers)",
    month = jul,
    year = "2025",
    url = "https://aclanthology.org/2025.acl-long.206/",
    doi = "10.18653/v1/2025.acl-long.206",
    pages = "4081--4108",
    ISBN = "979-8-89176-251-0"
}

@InProceedings{Carroll2019On,
author = {Carroll, Micah and Shah, Rohin and Ho, Mark and Griffiths, Tom and Seshia, Sanjit and Abbeel, Pieter and Dragan, Anca},
title = {On the utility of learning about humans for human-{AI} coordination},
pages={5174--5185},
year = {2019},
booktitle = {The 33rd International Conference on Neural Information Processing Systems},
articleno = {465},
numpages = {12}
}

@InProceedings{Jing2026Benchmarking,
  title = 	 {Benchmarking the Limits of In-Context Reinforcement Learning for Ad-Hoc Teamwork},
  author =       {Jing, Yuheng and Li, Kai and Zhang, Jiajun and Ma, Zeyao and Yang, Jiaxi and Zhang, Lei and Wu, Zhe and He, Jinmin and Xing, Junliang and Cheng, Jian},
  booktitle = 	 {International Conference on Machine Learning PMLR},
  pages = 	 {54278--54319},
  year = 	 {2026},
  editor = 	 {Zhang, Tong and Dudik, Miroslav and Jaggi, Martin and Agarwal, Alekh and Li, Sharon and Schuurmans, Dale and Zhu, Jerry and Berkenkamp, Felix and Dong, Hanze and Bietti, Alberto},
  volume = 	 {306},
  month = 	 {06--11 Jul},
  url = 	 {https://proceedings.mlr.press/v306/jing26d.html}
}

@article{Ruhdorfer2024The,
title={The {Overcooked} Generalisation Challenge: Evaluating Cooperation with Novel Partners in Unknown Environments Using Unsupervised Environment Design},
author={Constantin Ruhdorfer and Matteo Bortoletto and Anna Penzkofer and Andreas Bulling},
journal={Transactions on Machine Learning Research},
issn={2835-8856},
year={2025},
url={https://openreview.net/forum?id=K2KtcMlW6j},
note={}
}

@inproceedings{Liu2023LLM-powered,
  author       = {Jijia Liu and
                  Chao Yu and
                  Jiaxuan Gao and
                  Yuqing Xie and
                  Qingmin Liao and
                  Yi Wu and
                  Yu Wang},
  editor       = {Mehdi Dastani and
                  Jaime Sim{\~{a}}o Sichman and
                  Natasha Alechina and
                  Virginia Dignum},
  title        = {{LLM}-Powered Hierarchical Language Agent for Real-time Human-{AI} Coordination},
  booktitle    = {The 23rd International Conference on Autonomous Agents
                  and Multiagent Systems, {AAMAS} 2024, Auckland, New Zealand, May 6-10,
                  2024},
  pages        = {1219--1228},
  year         = {2024},
  url          = {https://dl.acm.org/doi/10.5555/3635637.3662979},
  doi          = {10.5555/3635637.3662979},
  bibsource    = {dblp computer science bibliography, https://dblp.org}
}

@inproceedings{Gessler2025OvercookedV2,
 author = {Gessler, Tobias and Dizdarevic, Tin and Calinescu, Ani and Ellis, Benjamin and Lupu, Andrei and Foerster, Jakob},
 booktitle = {International Conference on Learning Representations},
 editor = {Y. Yue and A. Garg and N. Peng and F. Sha and R. Yu},
 pages = {62752--62784},
 title = {{OvercookedV2}: Rethinking {Overcooked} for Zero-Shot Coordination},
 url = {https://proceedings.iclr.cc/paper_files/paper/2025/file/9d521f48cd3a9b218290a274af16905a-Paper-Conference.pdf},
 volume = {2025},
 year = {2025}
}

@inproceedings{Sun2025Collab-overcooked,
    title = "{Collab-Overcooked}: Benchmarking and Evaluating Large Language Models as Collaborative Agents",
    author = "Sun, Haochen  and
      Zhang, Shuwen  and
      Niu, Lujie  and
      Ren, Lei  and
      Xu, Hao  and
      Fu, Hao  and
      Zhao, Fangkun  and
      Yuan, Caixia  and
      Wang, Xiaojie",
    editor = "Christodoulopoulos, Christos  and
      Chakraborty, Tanmoy  and
      Rose, Carolyn  and
      Peng, Violet",
    booktitle = "The Conference on Empirical Methods in Natural Language Processing",
    month = nov,
    year = "2025",
    url = "https://aclanthology.org/2025.emnlp-main.249/",
    doi = "10.18653/v1/2025.emnlp-main.249",
    pages = "4922--4951",
    ISBN = "979-8-89176-332-6"
}

@inproceedings{Chaudhary2026Improving,
 author = {Chaudhary, Paresh and Liang, Yancheng and Chen, Daphne and Du, Simon and Jaques, Natasha},
 booktitle = {International Conference on Learning Representations},
 editor = {C. Vondrick and B. Hariharan and C. Raffel and L. Pinto and D. Yang and A. Faust},
 pages = {113415--113433},
 title = {Improving Human-{AI} Coordination through Online Adversarial Training and Generative Models},
 url = {https://proceedings.iclr.cc/paper_files/paper/2026/file/b837547467ea02deebdc9677d7f5a0be-Paper-Conference.pdf},
 volume = {2026},
 year = {2026}
}

@InProceedings{Ruhdorfer2025Unsupervised,
  title = 	 {Unsupervised Partner Design Enables Robust Ad-hoc Teamwork},
  author =       {Ruhdorfer, Constantin and Bortoletto, Matteo and Oei, Victor and Penzkofer, Anna and Bulling, Andreas},
  booktitle = 	 {International Conference on Machine Learning PMLR},
  pages = 	 {106186--106212},
  year = 	 {2026},
  editor = 	 {Zhang, Tong and Dudik, Miroslav and Jaggi, Martin and Agarwal, Alekh and Li, Sharon and Schuurmans, Dale and Zhu, Jerry and Berkenkamp, Felix and Dong, Hanze and Bietti, Alberto},
  volume = 	 {306},
  month = 	 {06--11 Jul},
  url = 	 {https://proceedings.mlr.press/v306/ruhdorfer26a.html}
}

@InProceedings{Wang2025CooT,
  title = 	 {{C}oo{T}: Learning to Coordinate In-Context with Coordination Transformers},
  author =       {Wang, Huai-Chih and Chuang, Hsiang-Chun and Cheng, Hsi-Chun and Wu, Dai-Jie and Sun, Shao-Hua},
  booktitle = 	 {International Conference on Machine Learning PMLR},
  pages = 	 {125587--125625},
  year = 	 {2026},
  editor = 	 {Zhang, Tong and Dudik, Miroslav and Jaggi, Martin and Agarwal, Alekh and Li, Sharon and Schuurmans, Dale and Zhu, Jerry and Berkenkamp, Felix and Dong, Hanze and Bietti, Alberto},
  volume = 	 {306},
  month = 	 {06--11 Jul},
  url = 	 {https://proceedings.mlr.press/v306/wang26ag.html}
}

@ARTICLE{Kim2026Co-Overcooked,
  title    = "{Co-Overcooked}: Cognitive Constraints on Partner Modeling in
              Human-{AI} Team Composition",
  author   = "Kim, Junghwan and Heo, Dongseok and Kim, Kyusik and Lee, Jung and
              Suh, Bongwon",
  journal  = "The Annual Meeting of the Cognitive Science Society",
  year     =  2026,
  volume   =  48,
}

@ARTICLE{Biswas2026Who,
  title     = "Who is helping whom? Analyzing inter-dependencies to evaluate
               cooperation in human-{AI} teaming",
  author    = "Biswas, Upasana and Palod, Vardhan and Bhambri, Siddhant and
               Kambhampati, Subbarao",
  journal   = "The AAAI Conference on Artificial Intelligence",
  publisher = "Association for the Advancement of Artificial Intelligence (AAAI)",
  volume    =  40,
  number    =  21,
  pages     = "17347--17356",
  month     =  mar,
  year      =  2026,
  language  = "en"
}

@inproceedings{Bortoletto2026ProToM,
    title = "{P}ro{T}o{M}: Promoting Prosocial Behaviour via Theory of Mind-Informed Feedback",
    author = "Bortoletto, Matteo  and
      Zhou, Yichao  and
      Ying, Lance  and
      Shu, Tianmin  and
      Bulling, Andreas",
    editor = "Liakata, Maria  and
      Moreira, Viviane P.  and
      Zhang, Jiajun  and
      Jurgens, David",
    booktitle = "Findings of the {A}ssociation for {C}omputational {L}inguistics",
    month = jul,
    year = "2026",
    url = "https://aclanthology.org/2026.findings-acl.1331/",
    doi = "10.18653/v1/2026.findings-acl.1331",
    pages = "26715--26734",
    ISBN = "979-8-89176-395-1"
}

@ARTICLE{Schroder2025Towards,
AUTHOR={Schr{\"o}der, Florian  and Heinrich, Fabian  and Kopp, Stefan },
TITLE={Towards fluid human-agent collaboration: From dynamic collaboration patterns to models of theory of mind reasoning},
JOURNAL={Frontiers in Robotics and AI},
VOLUME={Volume 12 - 2025},
YEAR={2025},
URL={https://www.frontiersin.org/journals/robotics-and-ai/articles/10.3389/frobt.2025.1532693},
DOI={10.3389/frobt.2025.1532693},
ISSN={2296-9144}}

@ARTICLE{Zhang2024ProAgent,
  title     = "{ProAgent}: Building proactive cooperative agents with large
               language models",
  author    = "Zhang, Ceyao and Yang, Kaijie and Hu, Siyi and Wang, Zihao and
               Li, Guanghe and Sun, Yihang and Zhang, Cheng and Zhang, Zhaowei
               and Liu, Anji and Zhu, Song-Chun and Chang, Xiaojun and Zhang,
               Junge and Yin, Feng and Liang, Yitao and Yang, Yaodong",
  journal   = "Conf. AAAI Artif. Intell.",
  publisher = "Association for the Advancement of Artificial Intelligence (AAAI)",
  volume    =  38,
  number    =  16,
  pages     = "17591--17599",
  month     =  mar,
  year      =  2024
}

@inproceedings{Kovac2023SocialAIschool,
  author = {Grgur Kova{\v{c}} and R{\'e}my Portelas and Peter Ford Dominey and Pierre-Yves Oudeyer},
  title = {{The SocialAI School: Insights from Developmental Psychology Towards Artificial Socio-Cultural Agents}},
  booktitle = {International Conference on Machine Learning First Workshop on Theory of Mind in Communicating Agents},
  year = {2023},
  url = {https://openreview.net/forum?id=HTrZnvD12V},
}

@misc{Agapiou2022Melting,
      title={Melting {Pot} 2.0}, 
      author={John P. Agapiou and Alexander Sasha Vezhnevets and Edgar A. Du{\'e}{\~n}ez-Guzm{\'a}n and Jayd Matyas and Yiran Mao and Peter Sunehag and Raphael K{\"o}ster and Udari Madhushani and Kavya Kopparapu and Ramona Comanescu and DJ Strouse and Michael B. Johanson and Sukhdeep Singh and Julia Haas and Igor Mordatch and Dean Mobbs and Joel Z. Leibo},
      year={2023},
      eprint={2211.13746},
      archivePrefix={arXiv},
      primaryClass={cs.MA},
      url={https://arxiv.org/abs/2211.13746}, 
}

@ARTICLE{Padmakumar2022TEACh,
  title    = "{TEACh}: Task-driven Embodied Agents that Chat",
  author   = "Padmakumar, Aishwarya and Thomason, Jesse and Shrivastava, Ayush
              and Lange, Patrick and Narayan-Chen, Anjali and Gella, Spandana
              and Piramuthu, Robinson and Tur, Gokhan and Hakkani-Tur, Dilek",
  journal  = "The AAAI Conference on Artificial Intelligence",
  pages    = "2017--2025",
  month    =  jun,
  year     =  2022
}

@inproceedings{Jain2020Acordialsync,
  title     = "A cordial sync: Going beyond marginal policies for multi-agent
               embodied tasks",
  author    = "Jain, Unnat and Weihs, Luca and Kolve, Eric and Farhadi, Ali and
               Lazebnik, Svetlana and Kembhavi, Aniruddha and Schwing, Alexander",
  year      =  2020,
  booktitle = "Computer Vision - ECCV 2020",
  pages     = "471--490",
  series    = "Lecture Notes in Computer Science",
}

@inproceedings{Park2023GenerativeAgents,
    apice = {GenagUist2023},
    author = {Park, Joon Sung and O'Brien, Joseph C. and Cai, Carrie Jun and Morris, Meredith Ringel and Liang, Percy and Bernstein, Michael S.},
    booktitle = {36th Annual ACM Symposium on User Interface Software and Technology (UIST'23)},
    doi = {10.1145/3586183.3606763},
    numpages = 22,
    openalex = {W4387835442},
    pages = {2:1--2:22},
    title = {Generative Agents: Interactive Simulacra of Human Behavior},
    url = {https://dl.acm.org/doi/10.1145/3586183.3606763},
    year = 2023
}

\newpage

\appendix

\counterwithin{table}{section}
\counterwithin{figure}{section}

\section{Agent MCP Servers}
\label{sec:agent_mcp}

EMIP uses Model Context Protocol (MCP) servers to connect agent control programs to the Minecraft environment and its audio communication functions. A separate instance of each of the Observation-Action, Voice, and Belief MCP servers is provided for each agent. When a single EMIP agent client controls multiple agents, it connects to the three server instances associated with each agent. Servers of the same type expose identical resource URIs and tool names; the client selects the target agent by connecting to that agent's server instance. This appendix describes the resources and tools exposed by each type of server.

\subsection{Observation-Action MCP}

The Observation-Action MCP server generates and provides structured observations from its associated agent's subjective perspective and supports physical action execution in Minecraft. The agent's name is configured using \texttt{setup\_minecraft\_interface} and stored in that server instance.

\paragraph{Retrieving Subjective Observations}

The server filters information received from Minecraft according to distance from the agent and occlusion by blocks to produce subjective observations. Through the two resources listed in Table~\ref{tab:observation-action-resources}, clients can retrieve the latest observation and observation management status in JSON format. In this appendix, each observation is associated with a tick value indicating logical time in Minecraft.

\begin{table*}[ht]
  \centering
  \small
  \renewcommand{\arraystretch}{1.15}
  \caption{Resources exposed by the Observation-Action MCP server.
  Each URI consists of \texttt{minecraft-observation://} followed by the name listed in the table.}
  \label{tab:observation-action-resources}
  \begin{tabularx}{\textwidth}{@{}lX@{}}
    \toprule
    \textbf{Resource} & \textbf{Information} \\
    \midrule
    \texttt{latest-observation}
    & The most recently received and processed subjective observation.\\
    \texttt{status}
    & The tick value of the latest observation, the initial tick value, the number of queued observations, the number of unprocessed Minecraft updates, the spatial bounds of the environment, the visibility distance, and the player name reserved for management. \\
    \bottomrule
  \end{tabularx}
\end{table*}

\paragraph{Action Execution and Observation Processing}

Clients invoke MCP tools to execute actions, process and retrieve observations, and generate prompts.
The server exposes the six tools listed in Table~\ref{tab:observation-action-tools}.

\begin{table*}[ht]
  \centering
  \small
  \renewcommand{\arraystretch}{1.15}
  \caption{Tools exposed by the Observation-Action MCP server and their functions.}
  \label{tab:observation-action-tools}
  \begin{tabularx}{\textwidth}{@{}lX@{}}
    \toprule
    \textbf{Tool} & \textbf{Function} \\
    \midrule
    \multicolumn{2}{@{}l}{\textbf{Agent Name Configuration}} \\
    \texttt{setup\_minecraft\_interface}
    & Sets the target agent's name and establishes its connection to Minecraft. \\
    \midrule
    \multicolumn{2}{@{}l}{\textbf{Action Execution}} \\
    \texttt{exec\_js}
    & Executes JavaScript code on the Mineflayer server to control the associated agent in Minecraft. \\
    \midrule
    \multicolumn{2}{@{}l}{\textbf{Observation Processing}} \\
    \texttt{start\_observation\_processing}
    & Starts or resumes processing of received Minecraft updates. \\
    \texttt{stop\_observation\_processing}
    & Stops observation processing. \\
    \midrule
    \multicolumn{2}{@{}l}{\textbf{Observation Retrieval and Prompt Generation}} \\
    \texttt{take\_queued\_observations}
    & Retrieves queued observations. If none are available, waits for a new observation. \\
    \texttt{render\_latest\_observation\_prompt}
    & Generates a prompt from the latest observation and an instruction. \\
    \bottomrule
  \end{tabularx}
\end{table*}

\subsection{Voice MCP}

The Voice MCP server supports audio communication between its associated agent and Minecraft. The client uses \texttt{easy\_llm\_voice\_setup} to send initial configuration information specifying the target agent and audio format to Minecraft. The server provides access to received audio data and speech recognition results, and sends synthesized speech or externally supplied audio to Minecraft. Each server instance separately maintains the agent's received audio buffer, speech recognition results, and read positions for incremental retrieval. The server exposes the eight tools listed in Table~\ref{tab:voice-tools} and no resources; audio data and speech recognition results are retrieved through tools.

\begin{table*}[ht]
  \centering
  \small
  \renewcommand{\arraystretch}{1.15}
  \caption{Tools exposed by the Voice MCP server and their functions.}
  \label{tab:voice-tools}
  \begin{tabularx}{\textwidth}{@{}lX@{}}
    \toprule
    \textbf{Tool} & \textbf{Function} \\
    \midrule
    \multicolumn{2}{@{}l}{\textbf{Audio Communication Configuration}} \\
    \texttt{easy\_llm\_voice\_setup}
    & Sends initial configuration information about the agent and audio format to Minecraft. \\
    \midrule
    \multicolumn{2}{@{}l}{\textbf{Audio and Speech Recognition Result Retrieval}} \\
    \texttt{get\_all\_audio}
    & Retrieves all data stored in the received audio buffer. \\
    \texttt{get\_new\_audio}
    & Retrieves audio buffer data added since the previous incremental retrieval and updates the read position. \\
    \texttt{get\_all\_asr\_result}
    & Retrieves all stored speech recognition results. \\
    \texttt{get\_new\_asr\_result}
    & Retrieves speech recognition results added since the previous incremental retrieval and updates the read position. \\
    \midrule
    \multicolumn{2}{@{}l}{\textbf{Audio Transmission Control}} \\
    \texttt{gen\_audio\_from\_text\_and\_send\_for\_minecraft}
    & Synthesizes speech from the specified text, converts it to Opus format, and sends it to Minecraft. \\
    \texttt{send\_pcm\_audio\_for\_minecraft}
    & Converts the supplied audio data to Opus format and queues it for sequential transmission to Minecraft. \\
    \texttt{stop\_pcm\_audio\_for\_minecraft}
    & Discards queued PCM audio data and sends an audio stop command to Minecraft after the chunk currently being processed has been transmitted. \\
    \bottomrule
  \end{tabularx}
\end{table*}

\subsection{Belief MCP}

The Belief MCP server manages and provides access to its associated agent’s subjective state and observation history. These are maintained using observations obtained through the Observation-Action MCP server and registered by the Harness.

\paragraph{Retrieving Subjective States and Observation Histories}

The server accumulates registered observations and uses them to update the agent's subjective state. Through the three resources listed in Table~\ref{tab:belief-resources}, clients can retrieve the current subjective state, stored observation history, and observation management status in JSON format.

\begin{table*}[ht]
  \centering
  \small
  \renewcommand{\arraystretch}{1.15}
  \caption{Resources exposed by the Belief MCP server.
  Each URI consists of \texttt{belief://} followed by the name listed in the table.}
  \label{tab:belief-resources}
  \begin{tabularx}{\textwidth}{@{}lX@{}}
    \toprule
    \textbf{Resource} & \textbf{Information} \\
    \midrule
    \texttt{current-state}
    & The current subjective state reflecting registered observations,
      including information about agents, blocks, and containers. \\
    \texttt{snapshot}
    & A snapshot containing the current subjective state, stored observation history and events, each observation's tick value, and the agent's name. \\
    \texttt{status}
    & The tick values of the latest registered observation and the current subjective state, together with the number of stored observations and their earliest and latest tick values. \\
    \bottomrule
  \end{tabularx}
\end{table*}

\paragraph{Information Management}

Clients use MCP tools to manage records, query observation histories and events, and generate prompts. The five tools listed in Table~\ref{tab:belief-tools} provide access to the controlled agent's subjective state and observation history, as well as estimates of other agents' subjective states and observation histories. These estimates are produced by the Harness from the controlled agent's own observations; they are not obtained by directly accessing other agents' internal states or histories.

\begin{table*}[ht]
  \centering
  \small
  \renewcommand{\arraystretch}{1.15}
  \caption{Tools exposed by the Belief MCP server and their functions.}
  \label{tab:belief-tools}
  \begin{tabularx}{\textwidth}{@{}lX@{}}
    \toprule
    \textbf{Tool} & \textbf{Function} \\
    \midrule
    \multicolumn{2}{@{}l}{\textbf{Record Management}} \\
    \texttt{add\_pending\_event}
    & Adds an event supplied by the agent to a queue for inclusion in the next observation registered. \\
    \midrule
    \multicolumn{2}{@{}l}{\textbf{History Queries}} \\
    \texttt{get\_observation\_history}
    & Retrieves stored observations within a specified tick range, returning the selected fields. \\
    \texttt{get\_events}
    & Retrieves stored events filtered by tick range, event type, and actor. \\
    \texttt{get\_state\_at}
    & Reconstructs and retrieves a past subjective state from information stored at or before the specified tick. \\
    \midrule
    \multicolumn{2}{@{}l}{\textbf{Prompt Generation}} \\
    \texttt{render\_prompt}
    & Generates a prompt from a template and variables using the current subjective state and stored history. \\
    \bottomrule
  \end{tabularx}
\end{table*}

\section{EMIP Manager Client}
\label{sec:manager_client}

The EMIP manager client provides the commands listed in Table~\ref{tab:manager-client} for configuring the experimental environment and controlling replay. These functions are separate from the agent control functions provided through EMIP agent clients.

\begin{table*}[ht]
  \centering
  \small
  \renewcommand{\arraystretch}{1.15}
  \caption{Functions of the EMIP manager client.}
  \label{tab:manager-client}
  \begin{tabularx}{\textwidth}{@{}lX@{}}
    \toprule
    \textbf{Command} & \textbf{Function} \\
    \midrule
    \multicolumn{2}{@{}l}{\textbf{Environment Management}} \\
    \texttt{speed} & Sets the rate at which the Minecraft environment advances and the observation interval. \\
    \texttt{save} & Saves the current environment state derived from the records to a file. \\
    \texttt{load-region} & Loads blocks from a saved Minecraft world into the current environment. \\
    \texttt{status} & Displays observation records, the environment's progression rate, and replay status. \\
    \texttt{/COMMAND} & Executes a Minecraft command with administrator privileges. \\
    \texttt{help} & Displays the available commands. \\
    \texttt{quit} & Exits the manager client, first restoring the environment to its pre-replay state if replay is in progress. \\
    \midrule
    \multicolumn{2}{@{}l}{\textbf{Replay Control}} \\
    \texttt{mark} & Marks the current time point as a named starting point for replay. \\
    \texttt{replay} & Replays recorded changes in the environment's state within the Minecraft world, beginning at a specified starting point. \\
    \texttt{pause} & Pauses replay. \\
    \texttt{resume} & Resumes paused replay. \\
    \texttt{restart} & Restarts replay from its starting point. \\
    \texttt{replay-speed} & Changes the replay speed. \\
    \texttt{end-replay} & Ends replay and restores the environment to its pre-replay state. \\
    \bottomrule
  \end{tabularx}
\end{table*}

\FloatBarrier

\section{Related Work}
\label{sec:related_work}

Table~\ref{tab:selected_prior_study-list} summarizes selected studies on cooperation and social interaction in simulated environments, together with related simulation platforms and evaluation benchmarks. In the Interaction column, AI-AI, H-H, and H-AI denote interactions among AI agents, among humans, and between humans and AI agents, respectively.

In the Platform column, M, O, H, V, and T denote Minecraft, Overcooked, Habitat, VirtualHome, and ThreeDWorld, respectively. Other denotes simulation platforms or evaluation environments outside these five categories. Multiple labels in either column indicate that the study covers more than one interaction type or platform.
The interaction types indicate those actually examined in the experiments or data collection of each study, rather than the full range of interactions supported by the respective platforms.
\vspace{10mm}

\setlength{\LTpre}{0pt}
\setlength{\LTpost}{0pt}
{\footnotesize
\setlength{\tabcolsep}{3pt}
\renewcommand{\arraystretch}{1.06}
\setlength{\LTpre}{0pt}
\setlength{\LTpost}{0pt}
\begin{longtable}{@{}>{\centering\arraybackslash}p{0.5cm} >{\RaggedRight\arraybackslash}p{9.0cm} >{\RaggedRight\arraybackslash}p{1.cm} >{\RaggedRight\arraybackslash}p{1.cm} >{\RaggedRight\arraybackslash}p{\dimexpr3.2cm+2\tabcolsep\relax}@{}}
\caption{Selected Prior Studies}\label{tab:selected_prior_study-list}\\
\toprule
\parbox[c][2.\baselineskip][c]{\linewidth}{\centering\textbf{No.}} & \parbox[c][2.\baselineskip][c]{\linewidth}{\textbf{Paper Title}} & \parbox[c][2.\baselineskip][c]{\linewidth}{\centering\textbf{Inter-}\\\textbf{action}} & \parbox[c][2.\baselineskip][c]{\linewidth}{\centering\textbf{Plat-}\\\textbf{form}} & \parbox[c][2.\baselineskip][c]{\linewidth}{\textbf{Venue and Date}} \\
\midrule
\endfirsthead
\multicolumn{5}{l}{\textit{Table \thetable\ (continued)}}\\
\toprule
\parbox[c][2.\baselineskip][c]{\linewidth}{\centering\textbf{No.}} & \parbox[c][2.\baselineskip][c]{\linewidth}{\textbf{Paper Title}} & \parbox[c][2.\baselineskip][c]{\linewidth}{\centering\textbf{Inter-}\\\textbf{action}} & \parbox[c][2.\baselineskip][c]{\linewidth}{\centering\textbf{Plat-}\\\textbf{form}} & \parbox[c][2.\baselineskip][c]{\linewidth}{\textbf{Venue and Date}} \\
\midrule
\endhead
\midrule
\multicolumn{5}{r}{\small Continued on next page}\\
\endfoot
\bottomrule
\endlastfoot
1 & Seeing Together: Multi-Robot Cooperative Egocentric Spatial Reasoning with Multimodal Large Language Models \citep{Peng2026Seeing} & AI-AI & H, Other & [1] arXiv; May 2026\newline [2] NeurIPS; Dec. 2026 (accepted)\\
\hline
2 & Lies We Can See: Joint Verbal and Non-Verbal Deception by VLM Agents in Embodied Social Interactions \citep{Ahn2026Lies} & AI-AI & M & [1] arXiv; Aug. 2026\newline [2] COLM Workshop on Agent Behavior; Oct. 2026 \\
\hline
3 & LLawCo: Learning Laws of Cooperation for Modeling Embodied Multi-Agent Behavior \citep{Zhou2026LLawCo} & AI-AI & H, T & [1] arXiv; Jun. 2026\newline [2] ICML; Jul. 2026 \\
\hline
4 & CollabBench: Benchmarking and Unleashing Collaborative Ability of LLMs with Diverse Players via Proactive Engagement \citep{Qian2026CollabBench} & AI-AI, H-AI & V, O & [1] arXiv; Jun. 2026\newline [2] ICML; Jul. 2026 \\
\hline
5 & Benchmarking the Limits of In-Context Reinforcement Learning for Ad-Hoc Teamwork \citep{Jing2026Benchmarking} & AI-AI & O & [1] arXiv; May 2026\newline [2] ICML; Jul. 2026 \\
\hline
6 & Unsupervised Partner Design Enables Robust Ad-hoc Teamwork \citep{Ruhdorfer2025Unsupervised} & AI-AI, H-AI & O & [1] arXiv; Aug. 2025\newline [2] ICML; Jul. 2026 \\
\hline
7 & CooT: Learning to Coordinate In-Context with Coordination Transformers \citep{Wang2025CooT} & AI-AI, H-AI & O & [1] arXiv; Jun. 2025\newline [2] ICML; Jul. 2026 \\
\hline
8 & Robots Ask the Way: Communication-Enabled Social Navigation \citep{Sacco2026Robots} & AI-AI & H & [1] arXiv; Jul. 2026\newline [2] IROS; Sep. 2026 \\
\hline
9 & Embodied Multi-Agent Coordination by Aligning World Models Through Dialogue \citep{Dongre2026Embodied} & AI-AI & H & [1] arXiv; May 2026\newline [2] SIGDIAL; Jul. 2026 \\
\hline
10 & Co-Overcooked: Cognitive Constraints on Partner Modeling in Human-AI Team Composition \citep{Kim2026Co-Overcooked} & AI-AI, H-AI, H-H & O & [1] CogSci; Jul. 2026 \\
\hline
11 & ProToM: Promoting Prosocial Behaviour via Theory of Mind-Informed Feedback \citep{Bortoletto2026ProToM} & AI-AI, H-AI, H-H & O & [1] arXiv; Sep. 2025\newline [2] Findings of ACL; Jul. 2026 \\
\hline
12 & MECoBench: A Systematic Study of Multimodal Agent Collaboration in Embodied Environments \citep{Liu2026MECoBench} & AI-AI & V & [1] arXiv; Jun. 2026 \\
\hline
13 & Multi-agent Framework for Time-Sensitive Complementary Collaboration in Minecraft \citep{Yi2026Multi-agent} & AI-AI & M & [1] arXiv; Jun. 2026 \\
\hline
14 & Collaborate and explain on-the-fly: knowledge-based reasoning and learning in ad hoc teamwork \citep{Dodampegama2026Collaborate} & AI-AI & V & [1] Frontiers in Artificial Intelligence; Jun. 2026 \\
\hline
15 & MindPower: Enabling Theory-of-Mind Reasoning in VLM-based Embodied Agents \citep{Zhang2026MindPower} & AI-AI & V, T & [1] arXiv; Nov. 2025\newline [2] CVPR; Jun. 2026 \\
\hline
16 & DRAMA: Next-Gen Dynamic Orchestration for Resilient Multi-Agent Ecosystems in Flux \citep{Zhao2026DRAMA} & AI-AI & V & [1] arXiv; Aug. 2025\newline [2] CVPR; Jun. 2026 \\
\hline
17 & SurveilNav: Collaborative Object Goal Navigation with Robot and Surveillance System \citep{Yu2026SurveilNav} & AI-AI & H & [1] arXiv; Jun. 2026\newline [2] ICRA; Jun. 2026 \\
\hline
18 & CommCP: Efficient Multi-Agent Coordination via LLM-Based Communication with Conformal Prediction \citep{Zhang2026CommCP} & AI-AI & H & [1] arXiv; Feb. 2026\newline [2] ICRA; Jun. 2026 \\
\hline
19 & Pragmatic Embodied Spoken Instruction Following in Human-Robot Collaboration with Theory of Mind \citep{Ying2026Pragmatic} & AI-AI, H-AI & V & [1] arXiv; Sep. 2024\newline [2] ICRA; Jun. 2026 \\
\hline
20 & Vulcan: Vision-Language-Model Enhanced Multi-Agent Cooperative Navigation for Indoor Fire-Disaster Response \citep{Liu2026VULCAN} & AI-AI & H & [1] arXiv; Apr. 2026\newline [2] IEEE INFOCOM EIN Workshop; May 2026 \\
\hline
21 & EnactToM: An Evolving Benchmark for Functional Theory of Mind in Embodied Agents \citep{Juneja2026EnactToM} & AI-AI & H & [1] arXiv; May 2026 \\
\hline
22 & MiTa: A Hierarchical Multi-Agent Collaboration Framework with Memory-Integrated and Task Allocation \citep{Zhang2026MiTa} & AI-AI & V & [1] arXiv; Jan. 2026\newline [2] ICASSP; May 2026 \\
\hline
23 & Gated Coordination for Efficient Multi-Agent Collaboration in Minecraft Game \citep{Jian2026Gated} & AI-AI & M & [1] arXiv; Apr. 2026 \\
\hline
24 & CoNavBench: Collaborative Long-Horizon Vision-Language Navigation Benchmark \citep{Wang2025CoNavBench} & AI-AI & H & [1] ICLR; Apr. 2026 \\
\hline
25 & From Assumptions to Actions: Turning LLM Reasoning into Uncertainty-Aware Planning for Embodied Agents \citep{Seo2026From} & AI-AI, H-AI & V, T & [1] arXiv; Feb. 2026\newline [2] ICLR; Apr. 2026 \\
\hline
26 & Improving Human-AI Coordination through Online Adversarial Training and Generative Models \citep{Chaudhary2026Improving} & AI-AI, H-AI & O & [1] arXiv; Apr. 2025\newline [2] ICLR; Apr. 2026 \\
\hline
27 & Won't, Not Can't: Designing a Refusal Agent for Long-Term Human–AI Collaboration \citep{Yamazaki2026Wont} & H-AI & M & [1] CHI EA; Apr. 2026 \\
\hline
28 & Co-NavGPT: Multirobot Cooperative Visual Semantic Navigation Using Vision Language Models \citep{Yu2026Co-NavGPT} & AI-AI & H & [1] arXiv; Oct. 2023\newline [2] IEEE Robotics and Automation Letters; Feb. 2026 \\
\hline
29 & Requesting Expert Reasoning: Augmenting LLM Agents with Learned Collaborative Intervention \citep{Wang2026Requesting} & H-AI & M & [1] arXiv; Feb. 2026 \\
\hline
30 & Embodied LLM Agents Learn to Cooperate in Organized Teams \citep{Guo2026Embodied} & AI-AI, H-AI & V & [1] arXiv; Mar. 2024\newline [2] NeurIPS Language Gamification Workshop; Dec. 2024\newline [3] IEEE Trans. on Computational Social Systems; Feb. 2026 \\
\hline
31 & Who Is Helping Whom? Analyzing Inter-Dependencies to Evaluate Cooperation in Human-AI Teaming \citep{Biswas2026Who} & AI-AI, H-AI & O & [1] arXiv; Feb. 2025\newline [2] AAAI; Jan. 2026 \\
\hline
32 & Improving the Efficiency of Collaboration Between Humans and Embodied AI Agents in 3D Virtual Environments \citep{Han2026Improving} & H-AI & M & [1] Applied Sciences; Jan. 2026 \\
\hline
33 & Investigating Feasibility of Large Language Model Agent Collaboration in Minecraft and Comparison with Human-Human Collaboration \citep{Hirota2025Investigating} & AI-AI, H-H & M & [1] IJCNLP-AACL; Dec. 2025 \\
\hline
34 & COOPERA: Continual Open-Ended Human-Robot Assistance \citep{Ma2025COOPERA} & AI-AI, H-AI & H & [1] arXiv; Oct. 2025\newline [2] NeurIPS; Dec. 2025 \\
\hline
35 & SoMi-ToM: Evaluating Multi-Perspective Theory of Mind in Embodied Social Interactions \citep{Fan2025SoMi-ToM} & AI-AI, H-AI & M & [1] arXiv; Jun. 2025\newline [2] NeurIPS Datasets and Benchmarks Track; Dec. 2025 \\
\hline
36 & MindForge: Empowering Embodied Agents with Theory of Mind for Lifelong Cultural Learning \citep{Lica2025MindForge} & AI-AI & M & [1] arXiv; Nov. 2024\newline [2] NeurIPS; Dec. 2025 \\
\hline
37 & Mind-Map Agent: Enhancing Cooperative Task Planning through Communication Alignment with Large Language Models \citep{Jeon2025Mind-Map} & AI-AI & V, T & [1] NeurIPS LAW Workshop; Dec. 2025 \\
\hline
38 & CausalMACE: Causality Empowered Multi-Agents in Minecraft Cooperative Tasks \citep{Chai2025CausalMACE} & AI-AI & M & [1] arXiv; Aug. 2025\newline [2] Findings of EMNLP; Nov. 2025 \\
\hline
39 & Collab-Overcooked: Benchmarking and Evaluating Large Language Models as Collaborative Agents \citep{Sun2025Collab-overcooked} & AI-AI, H-H & O & [1] arXiv; Feb. 2025\newline [2] EMNLP; Nov. 2025 \\
\hline
40 & Communicative PARTNR: Natural Language Communication under Partial Observability in Human–Robot Collaboration \citep{Jeon2025Communicative} & AI-AI & H & [1] ICCV Human-Robot-Scene Interaction and Collaboration Workshop; Oct. 2025 \\
\hline
41 & Collaborative Belief Reasoning with LLMs for Efficient Multi-Agent Collaboration \citep{Wang2025Collaborative} & AI-AI & V, T & [1] arXiv; Sep. 2025 \\
\hline
42 & Improving Cooperation in Collaborative Embodied AI \citep{Suprabha2026Improving} & AI-AI & V & [1] arXiv; Oct. 2025\newline [2] UKCI; Sep. 2025 \\
\hline
43 & Building an LLM-Based Teammate in Minecraft: A Testbed for Human-AI Collaboration \citep{Willett2025Building} & H-AI, H-H & M & [1] Human Factors and Ergonomics Society Annual Meeting; Sep. 2025 \\
\hline
44 & The Overcooked Generalisation Challenge: Evaluating Cooperation with Novel Partners in Unknown Environments Using Unsupervised Environment Design \citep{Ruhdorfer2024The} & AI-AI & O & [1] arXiv; Jun. 2024\newline [2] Transactions on Machine Learning Research; Sep. 2025 \\
\hline
45 & Acquiring complex knowledge and skills through digital simulation‐based training: Evidence from an agile project management teaching experience \citep{Coulon2026Acquiring} & H-H & M & [1] British Journal of Educational Technology; Aug. 2025 \\
\hline
46 & PillagerBench: Benchmarking LLM-Based Agents in Competitive Minecraft Team Environments \citep{Schipper2025PillagerBench} & AI-AI & M & [1] IEEE Conference on Games; Aug. 2025\newline [2] arXiv; Sep. 2025 \\
\hline
47 & Towards fluid human-agent collaboration: From dynamic collaboration patterns to models of theory of mind reasoning \citep{Schroder2025Towards} & H-H & O & [1] Frontiers in Robotics and AI; Aug. 2025 \\
\hline
48 & Leveraging Dual Process Theory in Language Agent Framework for Real-time Simultaneous Human-AI Collaboration \citep{Zhang2025Leveraging} & H-AI, AI-AI & O & [1] arXiv; Feb. 2025\newline [2] ACL 2025; Jul. 2025 \\
\hline
49 & Collaborative Tree Search for Enhancing Embodied Multi-Agent Collaboration \citep{Zu2025Collaborative} & H-AI, AI-AI & V, T & [1] CVPR; Jun. 2025 \\
\hline
50 & Learn as Individuals, Evolve as a Team: Multi-agent LLMs Adaptation in Embodied Environments \citep{Li2025Learn} & AI-AI & V, T & [1] arXiv; Jun. 2025 \\
\hline
51 & BeliefNest: A Joint Action Simulator for Embodied Agents with Theory of Mind \citep{Sagara2025BeliefNest} & AI-AI & M & [1] arXiv; May 2025 \\
\hline
52 & OvercookedV2: Rethinking Overcooked for Zero-Shot Coordination \citep{Gessler2025OvercookedV2} & AI-AI & O & [1] arXiv; Mar. 2025\newline [2] ICLR; Apr. 2025 \\
\hline
53 & CaPo: Cooperative Plan Optimization for Efficient Embodied Multi-Agent Cooperation \citep{Liu2025CaPo} & AI-AI, H-AI & V, T & [1] arXiv; Nov. 2024\newline [2] ICLR; Apr. 2025 \\
\hline
54 & PARTNR: A Benchmark for Planning and Reasoning in Embodied Multi-agent Tasks \citep{Chang2025PARTNR} & AI-AI, H-AI, H-H & H & [1] arXiv; Oct. 2024\newline [2] ICLR; Apr. 2025 \\
\hline
55 & EMOS: Embodiment-aware Heterogeneous Multi-robot Operating System with LLM Agents \citep{Chen2025EMOS} & AI-AI & H & [1] arXiv; Oct. 2024\newline [2] ICLR; Apr. 2025 \\
\hline
56 & Following the Human Thread in Social Navigation \citep{Scofano2025Following} & AI-AI & H & [1] arXiv; Apr. 2024\newline [2] ICLR; Apr. 2025 \\
\hline
57 & COMBO: Compositional World Models for Embodied Multi-Agent Cooperation \citep{Zhang2025COMBO} & AI-AI, H-AI & T & [1] arXiv; Apr. 2024\newline [2] ICLR; Apr. 2025 \\
\hline
58 & MultiCAT: Multimodal Communication Annotations for Teams \citep{Pyarelal2025MultiCAT} & H-H & M & [1] Findings of NAACL; Apr. 2025 \\
\hline
59 & Adaptive mechanisms of social and asocial learning in immersive collective foraging \citep{Wu2025Adaptive} & H-H & M & [1] bioRxiv; Jun. 2023\newline [2] Nature Communications; Apr. 2025 \\
\hline
60 & Collaborating Action by Action: A Multi-agent LLM Framework for Embodied Reasoning \citep{White2025Collaborating} & AI-AI & M & [1] arXiv; Apr. 2025 \\
\hline
61 & Enhancing Multi-Robot Semantic Navigation Through Multimodal Chain-of-Thought Score Collaboration \citep{Shen2025Enhancing} & AI-AI & H & [1] arXiv; Dec. 2024\newline [2] AAAI; Feb. 2025 \\
\hline
62 & REVECA: Adaptive Planning and Trajectory-Based Validation in Cooperative Language Agents Using Information Relevance and Relative Proximity \citep{Seo2025REVECA} & AI-AI, H-AI & V, T, O & [1] arXiv; May 2024\newline [2] AAAI; Feb. 2025 \\
\hline
63 & Artificial social intelligence in teamwork: how team traits influence human-AI dynamics in complex tasks \citep{Bendell2025Artificial} & H-AI, H-H & M & [1] Frontiers in Robotics and AI; Feb. 2025 \\
\hline
64 & TeamCraft: A Benchmark for Multi-Modal Multi-Agent Systems in Minecraft \citep{Long2024TeamCraft} & AI-AI & M & [1] arXiv; Dec. 2024 \\
\hline
65 & Project Sid: Many-agent simulations toward AI civilization \citep{Altera2024Project} & AI-AI & M & [1] arXiv; Oct. 2024 \\
\hline
66 & GOMA: Proactive Embodied Cooperative Communication via Goal-Oriented Mental Alignment \citep{Ying2024GOMA} & AI-AI, H-AI & V, O & [1] arXiv; Mar. 2024\newline [2] IROS; Oct. 2024 \\
\hline
67 & Collaborative Quest Completion with LLM-driven Non-Player Characters in Minecraft \citep{Rao2024Collaborative} & H-AI & M & [1] arXiv; Jul. 2024\newline [2] ACL Wordplay; Aug. 2024 \\
\hline
68 & VillagerAgent: A Graph-Based Multi-Agent Framework for Coordinating Complex Task Dependencies in Minecraft \citep{Dong2024VillagerAgent} & AI-AI & M, O & [1] arXiv; Jun. 2024\newline [2] Findings of ACL; Aug. 2024 \\
\hline
69 & MindAgent: Emergent Gaming Interaction \citep{Gong2024MindAgent} & AI-AI, H-AI & M, O & [1] arXiv; Sep. 2023\newline [2] Findings of NAACL; Jun. 2024 \\
\hline
70 & Habitat 3.0: A Co-Habitat for Humans, Avatars, and Robots \citep{Puig2024Habitat} & AI-AI, H-AI & H & [1] arXiv; Oct. 2023\newline [2] ICLR; May 2024 \\
\hline
71 & Building Cooperative Embodied Agents Modularly with Large Language Models \citep{Zhang2024Building} & AI-AI, H-AI & V, T & [1] arXiv; Jul. 2023\newline [2] ICLR; May 2024 \\
\hline
72 & LLM-Powered Hierarchical Language Agent for Real-time Human-AI Coordination \citep{Liu2023LLM-powered} & H-AI & O & [1] arXiv; Dec. 2023\newline [2] AAMAS; May 2024 \\
\hline
73 & MineLand: Simulating Large-Scale Multi-Agent Interactions with Limited Multimodal Senses and Physical Needs \citep{Yu2024MineLand} & AI-AI & M & [1] arXiv; Mar. 2024 \\
\hline
74 & ProAgent: Building proactive cooperative agents with large language models \citep{Zhang2024ProAgent} & AI-AI & O & [1] arXiv; Aug. 2023\newline [2] AAAI; Mar. 2024 \\
\hline
75 & Collective incentives reduce over-exploitation of social information in unconstrained human groups \citep{Deffner2024Collective} & H-H & Other & [1] Nature Communications; Mar. 2024\newline [2] PsyArXiv; Aug. 2023 \\
\hline
76 & Mitigating Generative Agent Social Dilemmas \citep{Yocum2023Mitigating} & AI-AI & M & [1] NeurIPS Workshop Foundation Models for Decision Making; Dec. 2023 \\
\hline
77 & Modeling Collaborative Dialogue in Minecraft with Action-Utterance Model \citep{Ichikawa2023Modeling} & H-H & M & [1] IJCNLP-AACL Student Research Workshop; Nov. 2023 \\
\hline
78 & Generative agents: Interactive simulacra of human behavior \citep{Park2023GenerativeAgents} & AI-AI & Other & [1] arXiv; Apr. 2023\newline [2] Annual ACM Symposium on User Interface Software and Technology; Oct. 2023 \\
\hline
79 & Adaptive Coordination in Social Embodied Rearrangement \citep{Szot2023Adaptive} & AI-AI & H & [1] arXiv; May 2023\newline [2] ICML; Jul. 2023 \\
\hline
80 & The SocialAI school: Insights from developmental psychology towards artificial Socio-cultural agents \citep{Kovac2023SocialAIschool} & AI-AI & Other & [1] ICML Workshop on Theory of Mind in Communicating Agents; Jul. 2023\newline [2] arXiv; Jul. 2023 \\
\hline
81 & Asynchronous Multi-Agent Reinforcement Learning for Efficient Real-Time Multi-Robot Cooperative Exploration \citep{Yu2023Asynchronous} & AI-AI & H & [1] arXiv; Jan. 2023\newline [2] AAMAS; May 2023 \\
\hline
82 & NOPA: Neurally-guided Online Probabilistic Assistance for Building Socially Intelligent Home Assistants \citep{Puig2023NOPA} & AI-AI, H-AI & V & [1] arXiv; Jan. 2023\newline [2] ICRA; May 2023 \\
\hline
83 & Melting Pot 2.0 \citep{Agapiou2022Melting} & AI-AI & Other & [1] arXiv; Nov. 2022 \\
\hline
84 & Learning Efficient Multi-agent Cooperative Visual Exploration \citep{Yu2022Learning} & AI-AI & H & [1] arXiv; Oct. 2021\newline [2] ECCV; Oct. 2022 \\
\hline
85 & Interactive Grounded Language Understanding in a Collaborative Environment: IGLU 2021 \citep{Kiseleva2022Interactive} & H-H & M & [1] NeurIPS Competitions and Demonstrations Track; Dec. 2021\newline [2] arXiv; May 2022 \\
\hline
86 & Craft an Iron Sword: Dynamically Generating Interactive Game Characters by Prompting Large Language Models Tuned on Code \citep{Volum2022Craft} & H-AI & M & [1] NAACL Wordplay; Jul. 2022 \\
\hline
87 & Analysis of Dialogue in Human-Human Collaboration in Minecraft \citep{Ichikawa2022Analysis} & H-H & M & [1] LREC; Jun. 2022 \\
\hline
88 & TEACh: Task-driven Embodied Agents that Chat \citep{Padmakumar2022TEACh} & AI-AI, H-AI, H-H & Other & [1] arXiv; Oct. 2021\newline [2] AAAI; Feb. 2022 \\
\hline
89 & The Utility of Explainable AI in Ad Hoc Human-Machine Teaming \citep{Paleja2022The} & H-AI & M & [1] arXiv; Sep. 2022\newline [2] NeurIPS; Dec. 2021 \\
\hline
90 & ThreeDWorld: A Platform for Interactive Multi-Modal Physical Simulation \citep{Gan2020ThreeDWorld} & AI-AI, H-AI & T & [1] arXiv; Jul. 2020\newline [2] NeurIPS Datasets and Benchmarks Track; Dec. 2021 \\
\hline
91 & MindCraft: Theory of Mind Modeling for Situated Dialogue in Collaborative Tasks \citep{Bara2021MindCraft} & H-H & M & [1] EMNLP; Nov. 2021 \\
\hline
92 & Watch-And-Help: A Challenge for Social Perception and Human-AI Collaboration \citep{Puig2020Watch-and-help} & AI-AI, H-AI & V & [1] arXiv; Oct. 2020\newline [2] ICLR; May 2021 \\
\hline
93 & A cordial sync: Going beyond marginal policies for multi-agent embodied tasks \citep{Jain2020Acordialsync} & AI-AI & Other & [1] arXiv; Jul. 2020\newline [2] ECCV; Oct. 2020 \\
\hline
94 & Understanding human-robot teams in light of all-human teams: Aspects of team interaction and shared cognition \citep{Demir2020Understanding} & H-H & M & [1] Human-Computer Studies; Aug. 2020 \\
\hline
95 & On the Utility of Learning about Humans for Human-AI Coordination \citep{Carroll2019On} & AI-AI, H-AI, H-H & O & [1] arXiv; Oct. 2019\newline [2] NeurIPS; Dec. 2019 \\
\hline
96 & Collaborative Dialogue in Minecraft \citep{Narayan-Chen2019Collaborative} & H-H & M & [1] ACL; Jul. 2019 \\
\hline
97 & CraftAssist: A Framework for Dialogue-enabled Interactive Agents \citep{Gray2019CraftAssist} & H-AI & M & [1] arXiv; Jul. 2019 \\
\hline
98 & The Multi-Agent Reinforcement Learning in MalmÖ (MARLÖ) Competition \citep{Perez-Liebana2019The} & AI-AI & M & [1] NeurIPS Workshop CiML; Dec. 2018\newline [2] arXiv; Jan. 2019 \\
\hline
99 & Investigating the role of Minecraft in educational learning environments \citep{Callaghan2016Investigating} & H-H & M & [1] Educational Media International; Nov. 2016 \\
\hline
100 & The Malmo Platform for Artificial Intelligence Experimentation \citep{Johnson2016The} & AI-AI, H-AI & M & [1] IJCAI Demonstration Track; Jul. 2016 \\
\hline
\end{longtable}
}

\section{Usage Examples}
\label{sec:example_english}

This appendix presents examples of interactions between a human-controlled (HC) agent and an AI agent in EMIP. In each figure, the horizontal axis, labeled Time (s), shows elapsed real time in seconds. The conversations were conducted in Japanese; the speech bubbles contain English translations of the original utterances. Green bands indicate periods of physical action, and wavy lines mark omitted time intervals.

\begin{sidewaysfigure}
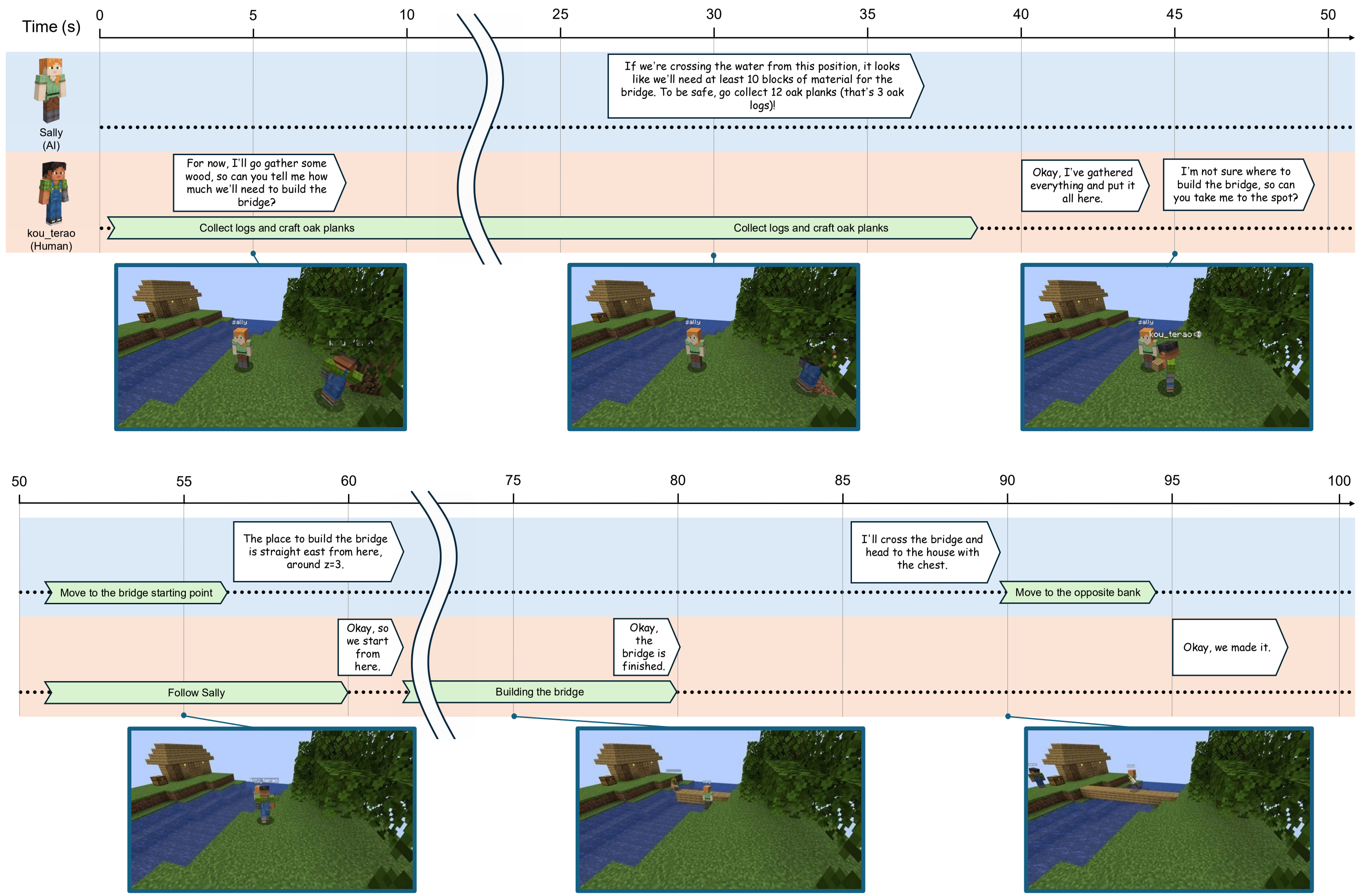

\centering
    \includegraphics[width=0.83\linewidth]{fig/App_small_ver_10240928/map2_start_10240928.pdf} \\
    \vspace{5mm}
    \includegraphics[width=0.83\linewidth]{fig/App_small_ver_10240928/map2_end_10240928.pdf}
    \caption{Collaborative bridge construction by an HC agent and an AI agent (English translations of utterances). While the HC agent collects logs and crafts planks, the AI agent explains how much material is needed for the bridge. The agents also move concurrently as the AI agent leads the way to the construction site and the HC agent follows. Speech bubbles show utterances, green bands indicate periods of physical action, and wavy lines mark omitted time intervals.}
    \label{fig:action1-english}
\end{sidewaysfigure}

Figure~\ref{fig:action1-english} illustrates concurrent spoken dialogue and physical actions, as well as concurrent physical actions by the two agents. While the HC agent collects logs and crafts planks, the AI agent explains how much material is needed to build a bridge. The agents also move concurrently as the AI agent leads the way to the construction site and the HC agent follows. This example illustrates how one participant can speak or act while the other is still performing an action.

\begin{sidewaysfigure}
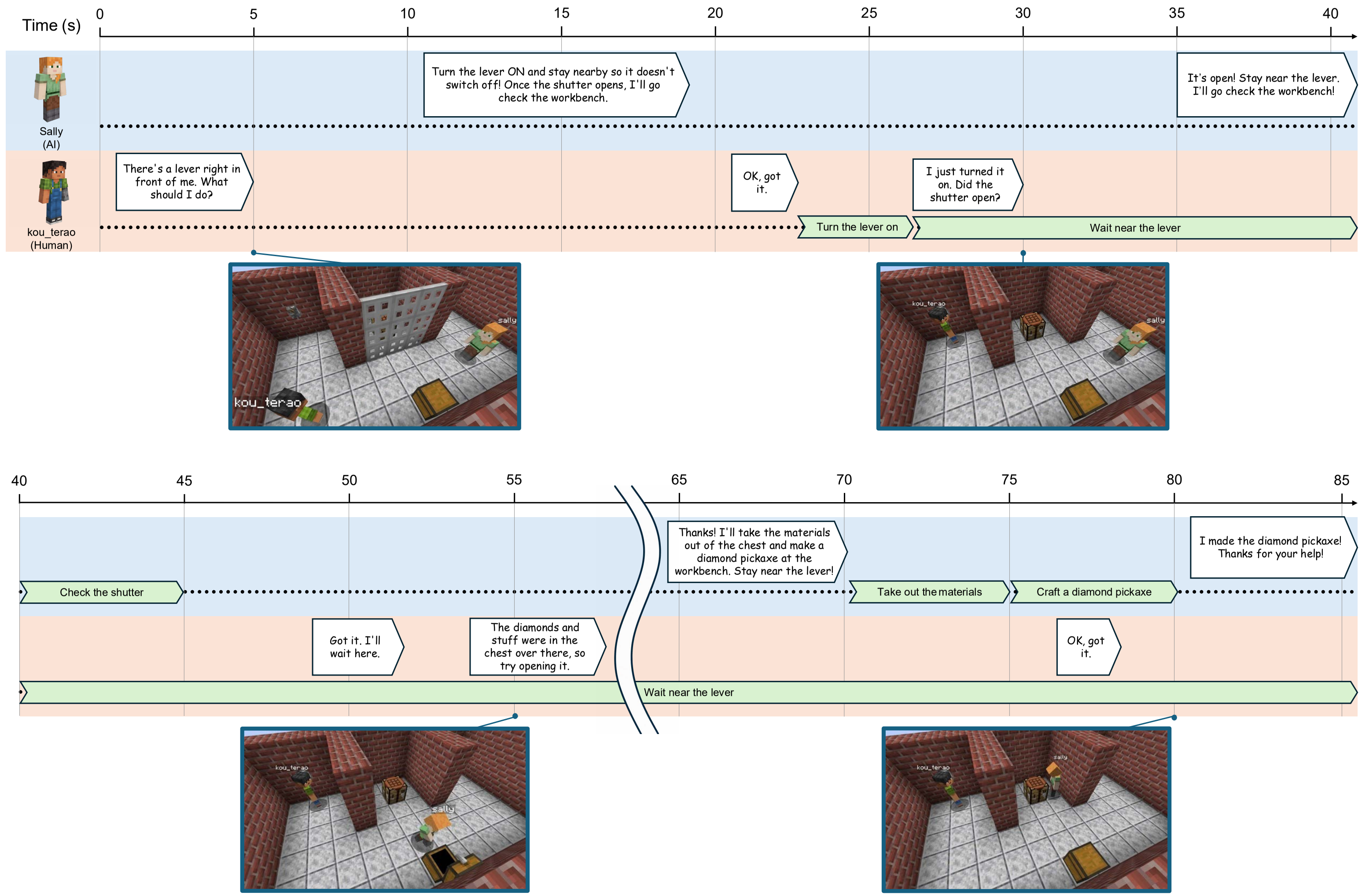

\centering
    \includegraphics[width=0.83\linewidth]{fig/App_small_ver_10240928/map4_start_10240928.pdf} \\
    \vspace{5mm}
    \includegraphics[width=0.83\linewidth]{fig/App_small_ver_10240928/map4_end_10240928.pdf}
    \caption{Joint task execution through role allocation and action coordination between an HC agent and an AI agent (English translations of utterances). The HC agent activates the lever and remains nearby to keep the shutter open while the AI agent checks the shutter, retrieves materials from a chest, and crafts a diamond pickaxe. The participants share task status through speech and coordinate their actions around the requirement for the HC agent to remain near the lever. Speech bubbles show utterances, green bands indicate periods of physical action, and wavy lines mark omitted time intervals.}
    \label{fig:action2-english}
\end{sidewaysfigure}

Figure~\ref{fig:action2-english} illustrates joint task execution through spoken information sharing and role allocation. The HC agent activates a lever and remains nearby to keep the shutter open. Meanwhile, the AI agent checks the shutter, retrieves materials from a chest, and crafts a diamond pickaxe. The HC agent's continued presence near the lever enables the AI agent to proceed with its work. The example illustrates how participants share task status through speech and coordinate their actions around this dependency.

\begin{sidewaysfigure}
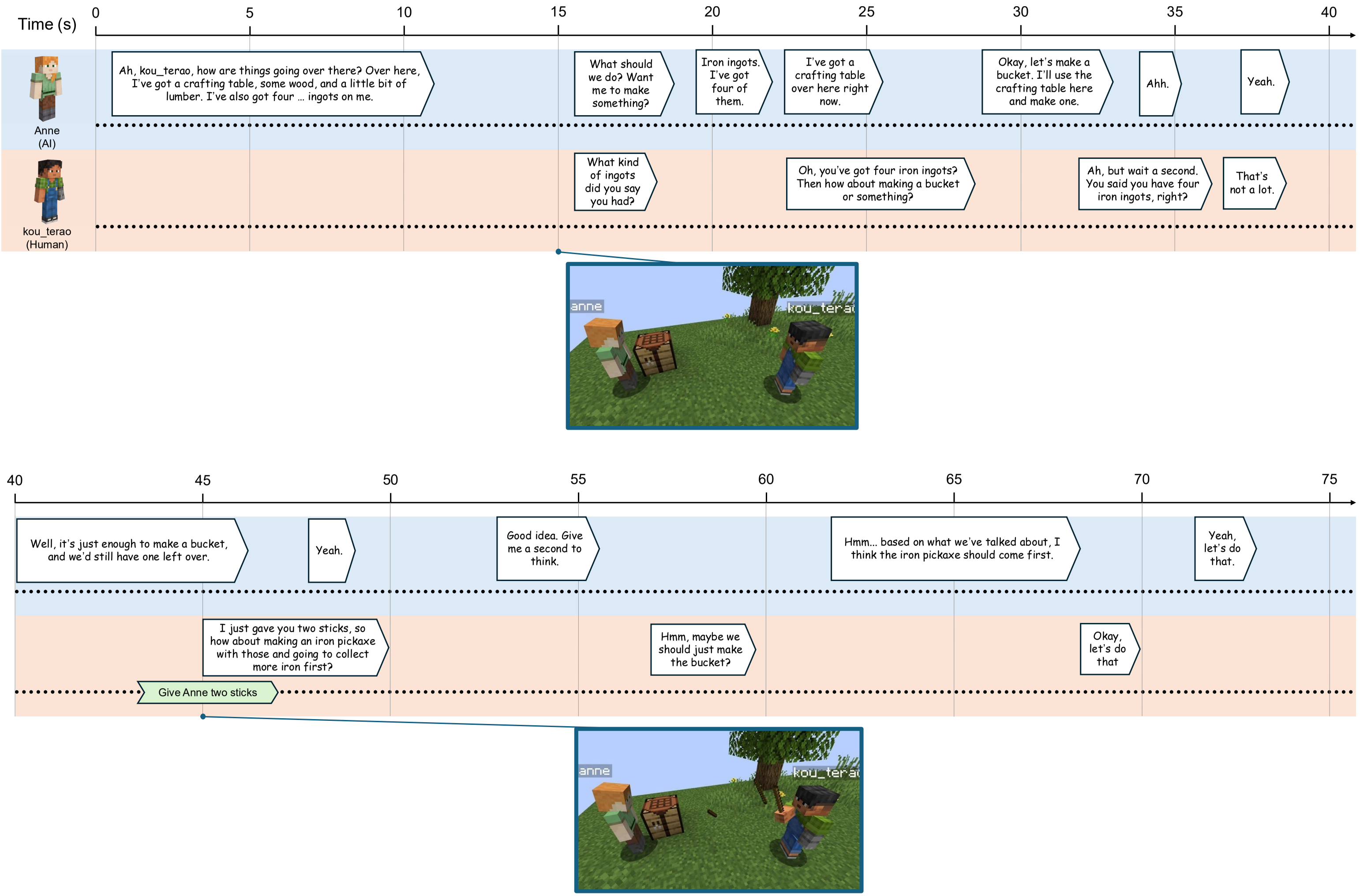

\centering
    \includegraphics[width=0.83\linewidth]{fig/App_small_ver_10240928/live_start_10240928.pdf} \\
    \vspace{5mm}
    \includegraphics[width=0.83\linewidth]{fig/App_small_ver_10240928/live_end_10240928.pdf}
    \caption{Full-duplex spoken dialogue with overlapping utterances between an HC agent and an AI agent (English translations of utterances). While discussing the available materials and their next task, the participants ask questions and provide backchannel responses during the other's speech. They consider crafting a bucket or an iron pickaxe and ultimately agree to craft the iron pickaxe first. Speech bubbles show utterances, and green bands indicate periods of physical action.}
    \label{fig:fullduplex-english}
\end{sidewaysfigure}

Figure~\ref{fig:fullduplex-english} illustrates full-duplex spoken dialogue with overlapping utterances. The HC agent and AI agent discuss the available materials and their next task. During this exchange, questions and backchannel responses overlap with the other participant's speech. They consider crafting a bucket or an iron pickaxe and ultimately agree to craft the iron pickaxe first.

\end{document}